\documentclass{article}

\usepackage[final,main]{neurips_2026}
\makeatletter
\renewcommand{\@noticestring}{}
\makeatother

\usepackage[utf8]{inputenc}
\usepackage[T1]{fontenc}
\usepackage{hyperref}
\usepackage{url}
\usepackage{booktabs}
\usepackage{amsfonts}
\usepackage{nicefrac}
\usepackage{microtype}
\usepackage{xcolor}
\usepackage{graphicx}
\usepackage{multirow}
\usepackage{amsmath}
\usepackage{amssymb}
\usepackage{listings}

\lstdefinestyle{mhar}{
  language=Python,
  basicstyle=\ttfamily\small,
  keywordstyle=\color[rgb]{0.0,0.5,0.0}\bfseries,
  stringstyle=\color[rgb]{0.75,0.0,0.0},
  commentstyle=\color[rgb]{0.5,0.5,0.5}\itshape,
  numberstyle=\tiny\color{gray},
  numbers=left,
  numbersep=6pt,
  xleftmargin=2em,
  framexleftmargin=1.5em,
  breaklines=true,
  showstringspaces=false,
  tabsize=4,
  columns=fullflexible,
  keepspaces=true,
  morekeywords={Tensor, Linear, RMSNorm, list, tuple, self, None},
}

\title{Multi-Head Attention Residuals}

\author{
  Cheng Luo\thanks{Equal contribution. Corresponding authors:
    \texttt{wdlctc@gmail.com}, \texttt{zefncai@gmail.com},
    \texttt{junjie.hu@wisc.edu}.} \\
  {\small Independent Researcher} \\
  {\small\texttt{wdlctc@gmail.com}} \\
  \And
  Zefan Cai$^*$ \\
  {\small University of Wisconsin--Madison} \\
  {\small\texttt{zefncai@gmail.com}} \\
  \And
  Junjie Hu \\
  {\small University of Wisconsin--Madison} \\
  {\small\texttt{junjie.hu@wisc.edu}} \\
}

\begin{document}

\maketitle

\begin{abstract}
Transformers propagate information across depth through a single additive
residual stream: every sublayer reads only the most recent state.
\emph{Attention residuals} relax this by letting each sublayer attend, through a
learned softmax. However, that read uses a \emph{single} query shared across the
entire width, so every feature subspace must read the depth history through one
distribution. The cost of this \emph{forced compromise} grows with how much
the subspaces disagree about which layers to read, and disagreement grows with
model width. We introduce \textbf{Multi-Head Attention Residuals (MHAR)}: the
routing query is reshaped into $H$ per-subspace heads, each with its own softmax
over the depth history. The read becomes block-diagonal, the reshape adds
\emph{zero} parameters and negligible compute, and $H{=}1$ recovers attention
residuals exactly. Trained from scratch on a deduplicated Nemotron-based
anneal corpus that is quality-filtered and STEM- and code-heavy, MHAR improves
validation loss over a standard Transformer at 100M, 350M, and 1B ($-0.061$,
$-0.149$, and $-0.140$). It achieves the best result among four
methods in every setting, with the gain increasing from 100M to the larger
scales. The head count is a real design axis rather than a free knob:
validation loss is U-shaped with respect to $H$, with a flat optimum at $H{=}4$
or $H{=}8$ across scales. We adopt $H{=}8$ for large-scale models;
over-splitting beyond this point ($H{=}16$) consistently gives back part of
the gain. A direct probe of the trained queries confirms that learned
subspace disagreement is the underlying driver. Fused Triton routing kernels
increase attention-residual training throughput from $0.2$--$0.5\times$ to
$0.55$--$0.88\times$ of the baseline while maintaining near-baseline peak
memory. An identity-preserving conversion using \emph{delta} attention
residuals supports 8B mid-training, yielding improvements of $+3.2$ on GSM8K
and $+3.1$ on GPQA.
\end{abstract}

\section{Introduction}
\label{sec:intro}

Transformers~\citep{vaswani2017attention} propagate information across depth
through a single residual stream~\citep{he2016deep}:
layer $\ell$ reads $\mathbf{h}_{\ell-1}$ and writes
$\mathbf{h}_\ell = \mathbf{h}_{\ell-1} + f_\ell(\mathbf{h}_{\ell-1})$. The
running sum couples every sublayer's input to the immediately preceding output,
even though useful features may have been computed many layers earlier.
\emph{Attention residuals} relax this by treating the set of all prior sublayer
outputs as a memory and letting each sublayer retrieve a learned,
softmax-weighted combination of them, an attention over depth rather than over
tokens~\citep{kimi2025attention}.

\begin{figure}[t]
\centering
\includegraphics[width=\textwidth]{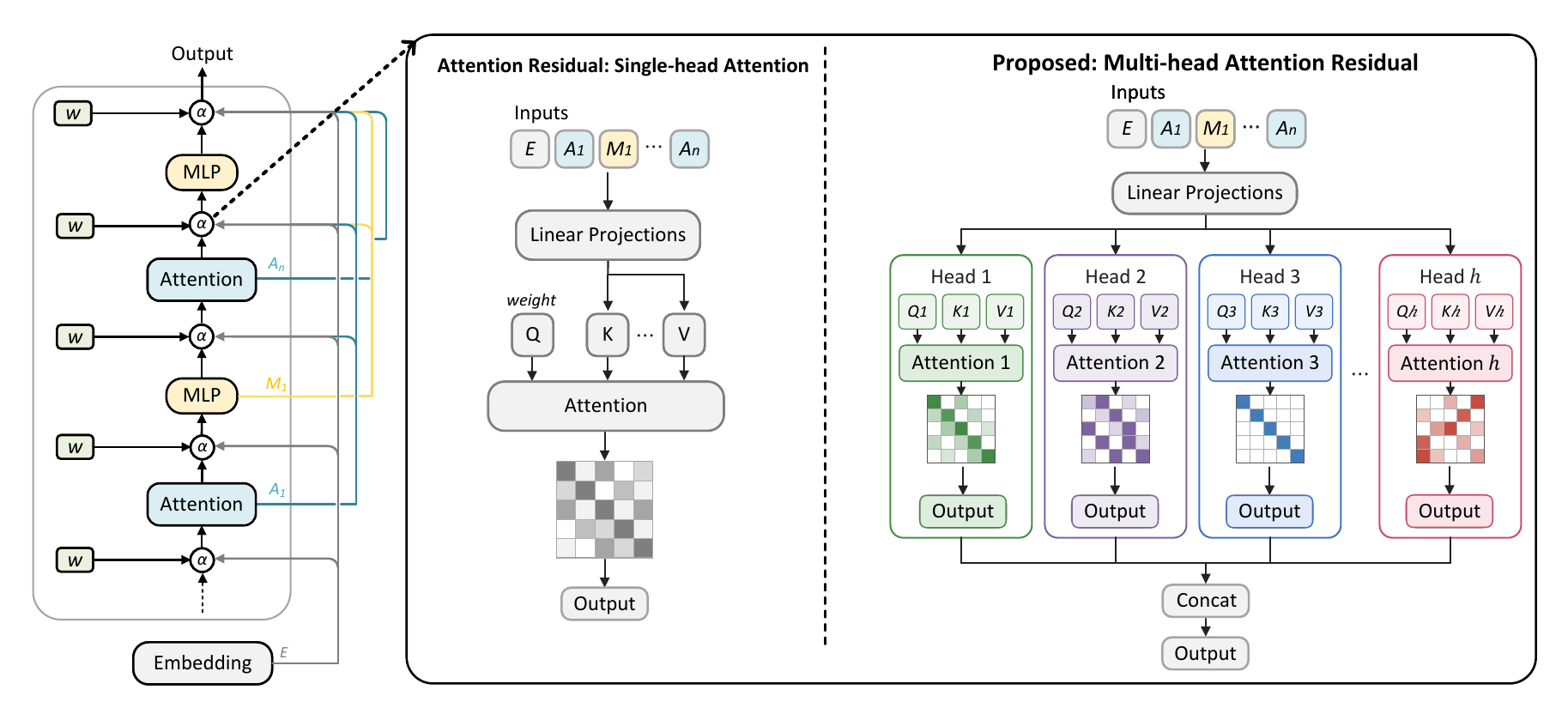}
\caption{\textbf{The depth read is an attention, so it should be multi-head.}
\textbf{Left:} attention residuals~\citep{kimi2025attention} feed every
sublayer a learned mixture of the depth history (embedding and every earlier
attention/MLP output). \textbf{Zoom left:} one routing site is exactly
\emph{single-head} attention over that history---one query, one depth map
shared by all $d$ channels (Eq.~\ref{eq:route}). \textbf{Zoom right:} MHAR
gives each of $H$ subspaces its own map (Eq.~\ref{eq:mh}), parameter-free;
$H{=}1$ recovers attention residuals exactly.}
\label{fig:teaser}
\end{figure}

This retrieval makes the depth history addressable, but only at the granularity of
the \emph{entire width}. A single learned query
$\mathbf{q}\in\mathbb{R}^d$~\citep{kimi2025attention} scores the $N{=}2L{+}1$ prior
sources and collapses them into one softmax distribution $\alpha$ over depth; all
$d$ coordinates of the retrieved mixture are then read through that same $\alpha$:
every feature dimension is forced to read from the same layers in the same
proportions (Figure~\ref{fig:teaser}, zoom left).

Structurally, this is a single self-attention query that routes over the depth
history rather than the token history. Self-attention is multi-head precisely
because one query cannot serve every subspace at once: the token read is already
split into heads; the depth read is not. The restriction is not benign: one shared
distribution collapses the differing depth preferences of every subspace into a
single read.

We therefore propose \textbf{Multi-Head Attention Residuals (MHAR)}, a strict
generalization that splits the routing query into $H$ per-subspace heads, each
routing independently over the depth history, and recovers attention residuals
exactly at $H{=}1$ (Figure~\ref{fig:teaser}, zoom right). The change is
parameter-free (the single $(d,)$ query is reshaped to $(H, d/H)$) and adds
negligible compute. The mechanism's real cost is at the systems level: every
sublayer re-reads the depth history, and reference implementations pay that
traffic several times over. We therefore also provide fused Triton routing
kernels that remove most of this overhead (Appendix~\ref{sec:kernel}), making
depth routing practical at scale.

We frame this restriction as a concrete cost, the \emph{forced compromise}: a
single query must reconcile, in one softmax, the differing depth preferences of
every subspace. Its cost is driven primarily by how much those subspaces
disagree, which grows with model width; a secondary, more speculative factor is
how collinear (hard to tell apart) the candidate sources are. Splitting the query
into per-subspace heads makes the read block-diagonal and removes the compromise.
We confirm the width dependence both in
loss (\S\ref{sec:exp}) and by directly probing the trained queries
(Appendix~\ref{sec:mechanism}).

Our contributions:
\begin{itemize}
\item \textbf{Multi-head attention residuals (MHAR).} We split the single
attention-residual routing query into $H$ per-subspace heads
(\S\ref{sec:method}): a reshape that adds no parameters or FLOPs, recovers
attention residuals exactly at $H{=}1$, and needs little tuning---the optimum
consistently lands at $H{=}4$ or $H{=}8$ across scales~\citep{ainslie2023gqa},
and for large-scale models we adopt $H{=}8$ (Appendix~\ref{app:kvh100}).
\item \textbf{Fused kernels for efficient MHAR.} Fused Triton routing kernels
lift attention-residual training from ${\sim}0.2$--$0.5\times$ to
$0.55$--$0.88\times$ baseline throughput at near-baseline peak memory
(Appendix~\ref{sec:kernel}), making depth routing practical at scale.
\item \textbf{Scaling MHAR.} Trained from scratch at 100M--1B, MHAR is the
best of four methods at every scale, with gains over the baseline that
\emph{grow} from 100M to the larger scales
($-0.061$/$-0.149$/$-0.140$; \S\ref{sec:exp}), and the
routing-head count shows a clear U-shaped optimum at $H{=}4$ or $H{=}8$
(we adopt $H{=}8$ at scale; \S\ref{sec:ablation}). An identity-preserving
conversion~\citep{luo2026delta} extends MHAR to 8B
mid-training, adding $+3.2$ GSM8K (paired $p{=}0.004$) over a schedule-matched
control (\S\ref{sec:midtrain}).
\end{itemize}

\section{Method: Multi-Head Attention Residuals}
\label{sec:method}

\begin{figure}[t]
\centering
\includegraphics[width=\textwidth]{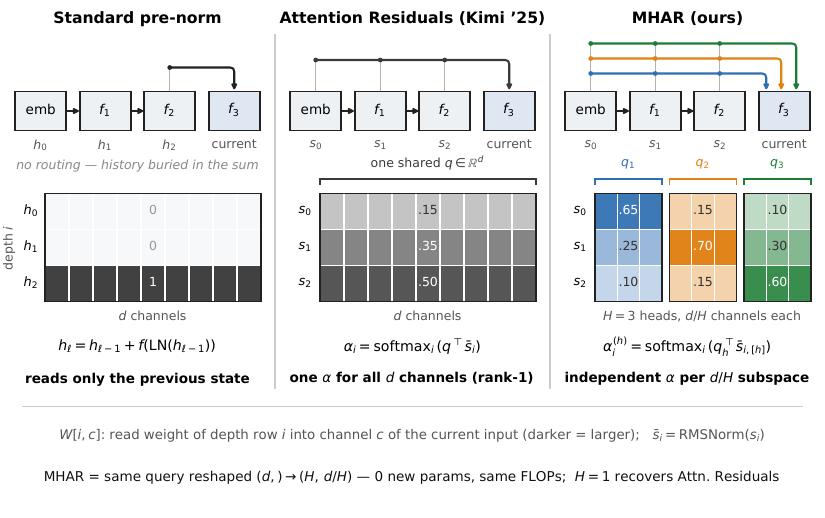}
\caption{\textbf{What each method lets the current sublayer read across depth.}
Each column: forward stack (top) and the depth-read weight matrix (bottom;
entry $[i,c]$ is how strongly channel $c$ reads depth row $i$; illustrative).
\textbf{Standard pre-norm:} a one-hot read of the previous state.
\textbf{Attention residuals:} one shared query reads all earlier sources---every
column identical (rank-1). \textbf{MHAR (ours):} each subspace reads depth
independently (block-wise columns).}
\label{fig:arch}
\end{figure}

\subsection{Background: the residual stream as depth memory}
A pre-norm decoder Transformer~\citep{vaswani2017attention,xiong2020prenorm} of
width $d$ with $L$ blocks maintains a single running state and updates it additively
(Figure~\ref{fig:arch}, left). Writing the two
sublayers of block $\ell$ (attention and MLP) as $f_\ell$ and $g_\ell$,
\begin{equation}
\mathbf{h}'_\ell = \mathbf{h}_{\ell-1} + f_\ell\big(\mathrm{LN}(\mathbf{h}_{\ell-1})\big),
\qquad
\mathbf{h}_\ell = \mathbf{h}'_\ell + g_\ell\big(\mathrm{LN}(\mathbf{h}'_\ell)\big).
\label{eq:resid}
\end{equation}
Because the state is a running sum, every sublayer reads only the most recent
state $\mathbf{h}_{\ell-1}$. Features computed many layers earlier are still
present in the sum, but are no longer individually addressable: the network
cannot selectively re-read the output of, say, layer~3 when computing layer~30
without that information first surviving every intervening additive update.

\begin{figure}[t]
\begin{lstlisting}[style=mhar]
def route(self, sources: list[Tensor], proj: Linear, norm: RMSNorm):
    """Multi-head depth routing: split the (D,) query into self.H heads
       and run H independent softmaxes over the sources. This kernel is
       the ONLY change from attention residuals (Kimi 2025)."""
    V = torch.stack(sources)                   # [N, B, T, D]
    N, B, T, D = V.shape
    q = proj.weight.view(self.H, D // self.H)  # (D,) query -> H heads
    Vh = V.view(N, B, T, self.H, D // self.H)  # H slices per source
    logits = einsum('h e, n b t h e -> n b t h', q, norm(Vh))
    alpha = logits.softmax(0)                  # H softmaxes over depth
    mixed = einsum('n b t h, n b t h e -> b t h e', alpha, Vh)
    return mixed.reshape(B, T, D)

def forward(self, sources):  # IDENTICAL to attention residuals (Kimi 2025)
    # --- Attention sublayer ---
    h = self.route(sources, self.attn_proj, self.attn_norm)
    sources.append(self.attn(norm(h)))   # append output as a new source

    # --- MLP sublayer ---
    h = self.route(sources, self.mlp_proj, self.mlp_norm)
    sources.append(self.mlp(norm(h)))    # replacement: append, don't add

    return sources
\end{lstlisting}
\caption{\textbf{Multi-Head Attention Residuals pseudocode.} The \texttt{forward} pass is
\emph{identical} to attention residuals~\citep{kimi2025attention}: each sublayer reads a
routed combination of the prior sources and \emph{appends} its raw output as a new source
(replacement routing). The \emph{sole} change is the \texttt{route} kernel---it splits the
single $(d,)$ query into \texttt{self.H} heads and runs $H$ independent softmaxes over
depth, adding \emph{zero} parameters. Substituting the single-head kernel ($H{=}1$)
recovers attention residuals exactly (Appendix~\ref{app:singlehead}).}
\label{fig:code}
\end{figure}

\subsection{Attention residuals}
\emph{Attention residuals}~\citep{kimi2025attention} make past sublayer outputs
individually addressable by treating them as a memory and \emph{routing} over
them. Let the network expose the ordered list of \emph{sources}
$\mathcal{S} = (\mathbf{s}_0, \mathbf{s}_1, \dots, \mathbf{s}_{N-1})$ of $N{=}2L{+}1$
sources, where $\mathbf{s}_0$ is the token embedding and each later source is the raw
output of one attention or MLP sublayer. In place of the additive update
(Eq.~\ref{eq:resid}), the input to a sublayer is a learned, softmax-weighted
combination of \emph{all} sources produced so far. With a per-sublayer learned
pseudo-query $\mathbf{q}\in\mathbb{R}^{d}$ and an RMSNorm applied to the sources
as keys,
\begin{equation}
\alpha_i = \frac{\exp\!\big(\mathbf{q}^\top \mathrm{RMSNorm}(\mathbf{s}_i)\big)}
{\sum_{j=0}^{M}\exp\!\big(\mathbf{q}^\top \mathrm{RMSNorm}(\mathbf{s}_j)\big)},
\qquad
\tilde{\mathbf{h}} = \sum_{i=0}^{M} \alpha_i\,\mathbf{s}_i ,
\label{eq:route}
\end{equation}
where $M$ indexes the sources available at that point. This is an attention over
\emph{depth} (the source/layer axis) rather than over tokens; the analogue of
keys are the normalized sources and the analogue of values are the sources
themselves, with a single learned query per routing site (Figure~\ref{fig:arch}, middle).

\subsection{Multi-head routing}
A single query (Eq.~\ref{eq:route}) forces \emph{all} $d$ coordinates of the
routed mixture to share one distribution $\alpha$ over depth: every feature
dimension must read from the same layers in the same proportions. Different
feature subspaces, however, may be best served by different layers. We remove
this constraint with \textbf{multi-head routing (MHAR)} (Figure~\ref{fig:arch}, right):
partition the
coordinates into $H$ contiguous heads of width $d/H$, give each head its own
query, and let each head route independently. For head $h$ with
query $\mathbf{q}_h\in\mathbb{R}^{d/H}$ and writing $\mathbf{x}_{[h]}$ for the
$h$-th slice of a vector $\mathbf{x}$,
\begin{equation}
\alpha^{(h)}_i = \mathrm{softmax}_i\!\Big(\mathbf{q}_h^\top\,\mathrm{RMSNorm}(\mathbf{s}_i)_{[h]}\Big),
\qquad
\tilde{\mathbf{h}}_{[h]} = \sum_{i} \alpha^{(h)}_i\,\mathbf{s}_{i,[h]},
\label{eq:mh}
\end{equation}
and the $H$ routed slices are concatenated to width $d$. Each head scores only
its own slice of each source and mixes only that slice, so the $H$ softmaxes are
fully independent and different subspaces can attend to different layers.
Figure~\ref{fig:code} shows the implementation: the \texttt{forward} pass is
\emph{identical} to attention residuals~\citep{kimi2025attention}; the \emph{only} change
is the \texttt{route} kernel (the single-head original is recovered at $H{=}1$,
Appendix~\ref{app:singlehead}).

\paragraph{Properties.}
\emph{(i) Parameter-matched.} The $H$ queries
$\{\mathbf{q}_h\}_{h=1}^{H}$ together form an $(H, d/H)$ tensor with exactly $d$
parameters---the same as the single $(d,)$ query. Multi-head routing therefore
adds \emph{zero} parameters over single-head routing; $H$ is a reshape, not a
widening.
\emph{(ii) FLOP-matched.} The scoring and mixing in Eq.~\ref{eq:mh} touch each
source coordinate exactly once, as in Eq.~\ref{eq:route}; the only overhead is
computing $H$ small softmaxes over the depth axis (length $\le 2L{+}1$) instead
of one, which is negligible next to the attention and MLP sublayers.
\emph{(iii) Strict generalization of attention residuals.} At $H{=}1$ the route
reduces to the single shared query of Eq.~\ref{eq:route} run through the identical
forward, so the \emph{single-head} method in our tables \emph{is} the original
attention-residual formulation of~\citet{kimi2025attention}
(Appendix~\ref{app:singlehead})---making MHAR-vs-single-head a strict
iso-parameter, iso-forward control.

\paragraph{Fused routing kernels.} Depth routing is memory-bound: scoring and mixing (Eq.~\ref{eq:mh}) perform only a constant number of operations per source element read, with none of the data reuse that lets the attention and MLP matrix multiplies run compute-bound, so its cost is the bytes it moves, and every sublayer re-reads the whole depth history ($O(L^2\,B\,T\,d)$ traffic per step). A reference implementation pays this traffic several times over and retains the per-sublayer stacked sources, the dominant activation cost. We provide fused, deterministic Triton forward/backward kernels that cut per-microbatch routing time by $2\times$ over \texttt{torch.compile} and $6.4\times$ over the eager reference (${\sim}30\%$ end-to-end mid-training throughput on $8\times$H100). Appendix~\ref{sec:kernel} gives the full forward/backward algorithm, the delta variant used for mid-training, and numerical verification.

\section{Experiments}
\label{sec:exp}

\subsection{Setup}

\paragraph{Models and training.} We train decoder-only Transformers from scratch on
the \texttt{anneal\_pt\_v3} corpus---a deduplicated, Nemotron-based anneal
mix~\citep{nvidia2025nemotronccv2} (quality-filtered, synthetic-, STEM-, and
code-heavy), the same corpus used for the 8B mid-training experiments of
\S\ref{sec:midtrain} (composition in Appendix Table~\ref{tab:datamix})---for
20K steps with the AdamW
optimizer~\citep{loshchilov2019adamw}, a cosine learning-rate schedule with
1{,}000 warmup steps decaying to $0.1\times$ peak, and bfloat16. (A replication of the same
four-method comparison on a second, web-generic corpus is
reported in Appendix Table~\ref{tab:mainweb}.) The three scales
are: 100M ($d{=}512$, $L{=}12$, 8 heads, 4 KV heads, FFN 1536), 350M
($d{=}1024$, $L{=}24$, 16 heads, 8 KV heads, FFN 4096), and 1B ($d{=}1280$,
$L{=}36$, 16 heads, 8 KV heads, FFN 5120), all with head dimension 64--80 and a
Qwen3-style architecture~\citep{qwen2025qwen3}. The 100M models use sequence
length 2048 and global batch 64; the 350M/1B models use sequence length 1024 and
global batch 32. We compare four methods at each scale: a standard Transformer
(\emph{baseline}), hyper-connections~\citep{zhu2024hyperconnections}, single-head
attention residuals ($H{=}1$, i.e.\ the published attention-residual
formulation of~\citet{kimi2025attention}), and multi-head attention residuals
(\emph{MHAR}) with the number of routing heads $H$ set equal to the number of
KV heads. Following~\citet{kimi2025attention}, all routing pseudo-queries are
\emph{zero-initialized}, so every depth softmax starts as a uniform average
over its sources (the appendix replication predates this fix and uses randomly
initialized queries; see the caveat there). All four methods share architecture,
data, schedule, optimizer, and data-parallel degree,
so the routing mechanism is the only difference. Table~\ref{tab:main} reports all
methods at a unified peak learning rate of $5\times10^{-4}$; a per-method
learning-rate sweep (Appendix Table~\ref{tab:lrsweep}) shows the
method ranking is not an artifact of sharing one rate. MHAR is
parameter-matched to baseline up to the $O(d)$ routing queries and adds only
$0.5$--$1.2\%$ routing FLOPs (Table~\ref{tab:maincost}). Full training and reproducibility details (optimizer
hyperparameters, tokenizer, positional scheme, precision, and the validation
protocol) are given in Appendix~\ref{app:repro}.

\paragraph{Metric.} A single validation pass is noisy at 1B ($\sim\!\pm0.07$ over an
evaluation pass; Appendix~\ref{sec:mechanism}), so unless noted we report the
\emph{tail-mean} of the last eleven evaluations (steps 15--20k). This $\pm0.07$ is
single-pass sampling noise; the cross-method $\Delta$ is far better resolved,
because methods are compared \emph{paired} (same data order within a scale) and
averaged over eleven evals. Pairing matters doubly here: each streaming
validation pass samples a small window of the corpus, and long documents make
per-eval values fluctuate across steps, so only same-step paired deltas and
tail-means are meaningful. The
single-cell tables report one fixed seed (Appendix~\ref{app:repro}).

\subsection{Pretraining}

\begin{table}[h]
\centering
\caption{From-scratch validation loss ($\downarrow$) after a unified 20K-step
schedule on the \texttt{anneal\_pt\_v3} corpus at a unified peak learning rate
of $5\times10^{-4}$ (tail-mean over the last quarter of training;
\S\ref{sec:exp}); $\Delta$ is relative
to the same-rate baseline. MHAR is best at every scale, and both
attention-residual variants \emph{grow} their gain from 100M to the larger
scales. The last
column is an \emph{over-splitting control}: the same model with the routing
query split into $H{=}16$ heads, past the $H{=}4$--$8$ optimum. A replication on a
second corpus, where the single-head column reverses, is in Appendix
Table~\ref{tab:mainweb}.}
\label{tab:main}
\begin{tabular}{lcccccc}
\toprule
\textbf{Scale} & \textbf{$d$/$L$/KV} & \textbf{Baseline} & \textbf{Hyper-conn.} & \textbf{Single-head} & \textbf{MHAR ($H{=}8$)} & \textbf{MHAR ($H{=}16$)} \\
\midrule
100M & 512/12/4  & 3.031 & 2.999 \,($-0.032$) & 2.970 \,($-0.060$) & \textbf{2.969} \,($-0.061$) & 2.996 \,($-0.035$) \\
350M & 1024/24/8 & 2.997 & 2.931 \,($-0.066$) & 2.876 \,($-0.121$) & \textbf{2.848} \,($-0.149$) & 2.895 \,($-0.102$) \\
1B   & 1280/36/8 & 2.894 & 2.807 \,($-0.086$) & 2.759 \,($-0.134$) & \textbf{2.754} \,($-0.140$) & 2.825 \,($-0.069$) \\
\bottomrule
\end{tabular}
\end{table}

\paragraph{Comparison to hyper-connections.} The most relevant learned alternative
to the additive residual is hyper-connections~\citep{zhu2024hyperconnections}, which
replace the single stream with $n$ parallel streams mixed by per-layer learned depth
and width connections. Trained in the same sweep ($n{=}4$), they improve over the
baseline at all three scales ($-0.032$/$-0.066$/$-0.086$; Table~\ref{tab:main}),
but \textbf{MHAR outperforms hyper-connections at every scale} (by $0.029$,
$0.083$, and $0.053$ at 100M/350M/1B): enriching the
depth \emph{routing} of a single stream is consistently stronger than enriching
the \emph{connection topology} across streams, at a fraction of the parameter
and memory-traffic cost (Table~\ref{tab:maincost}).

\paragraph{Over-splitting control ($H{=}16$).} The head count is bounded on
both sides. The last column of Table~\ref{tab:main} repeats MHAR with the same
$(d,)$ query reshaped into $H{=}16$ heads---\emph{past} the $H{=}4$--$8$ optimum
($2\times$ the KV count at 350M/1B, $4\times$ at 100M), every other setting identical.
At 350M this gives back a third of MHAR's gain ($-0.149\rightarrow-0.102$)
and at 1B half of it ($-0.140\rightarrow-0.069$); at 100M the penalty is
present but shallower. At all three scales $H{=}16$ lands \emph{above}
single-head routing. The
over-splitting cost grows monotonically with scale ($+0.026$/$+0.047$/$+0.071$
over the $H{=}4$--$8$ optimum), mirroring how the single-query compromise grows with width: each head now routes a
subspace narrower than what any KV group consumes, so coherent features are
split across independently-routed slices. Validation loss is therefore
\emph{U-shaped} in $H$, bottoming out in a flat basin at $H{=}4$--$8$---head
count is a real
design axis, and $H{=}4$ or $H{=}8$ is its near-optimal setting across scales
(we adopt $H{=}8$ at scale; full sweep in \S\ref{sec:ablation}).

\paragraph{Compute-equivalent gain.} The loss deltas of
Table~\ref{tab:main} can also be expressed as compute-equivalent
gain~\citep{davidson2023ceg,mai2026thinking}: the factor by which baseline
training FLOPs would have to grow to match MHAR's loss, under a power law
$L{=}AC^{-\alpha}$ fitted to the lower envelope of the three baseline
training curves in the style of~\citet{kaplan2020scaling}
($\alpha{=}0.089$; the cosine schedule makes mid-run points pessimistic, so
this envelope is a \emph{conservative} exponent). Because MHAR's routing adds
only $0.5$--$1.2\%$ FLOPs, the loss gain translates into a
$1.3\times$/$1.8\times$/$1.7\times$ compute equivalence at 100M/350M/1B. The exponent brackets the estimate:
fitting only the three final points gives $\alpha{=}0.053$---nearly identical
to the $0.057$ that \citet{kimi2025attention} fit on their own ladder---and
would give $1.5$--$2.6\times$, while a steeper Chinchilla-like
$\alpha{=}0.15$ gives $1.15$--$1.4\times$.

Beyond the final numbers, MHAR dominates the baseline \emph{throughout}
training, not just at convergence: its loss curve stays below the baseline from
early in training onward (Figure~\ref{fig:trainloss}, Appendix~\ref{app:robustness};
the 100M web-replication pair shown). Since the two runs share
node, software, data order, and global batch, the separation is attributable to
the routing mechanism alone.

\begin{figure}[t]
\centering
\includegraphics[width=\textwidth]{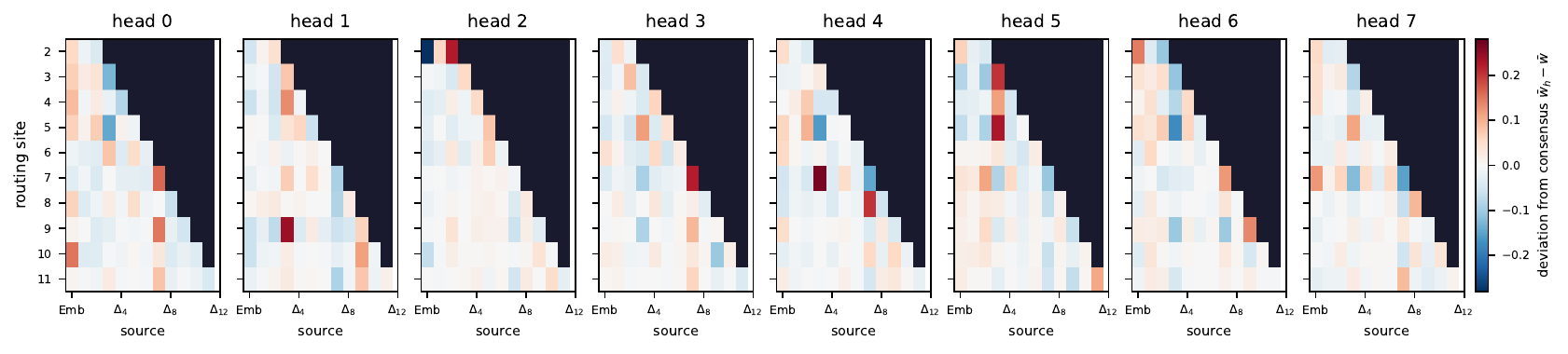}
\caption{\textbf{The $H$ heads carry genuinely different depth-links.} Each
head's token-averaged deviation $\bar{w}_h-\bar{w}$ from the head-consensus
routing at the first ten routing sites of the trained 1B MHAR model (dark
cells: source not yet available). Deviations reach $\pm0.28$ vs.\ $0.067$
under a matched-norm random query, replicate on disjoint evaluation text
($r{=}0.77$), and are near-uncorrelated across heads.}
\label{fig:perhead}
\end{figure}

\paragraph{When do the extra heads matter?} A single routing query shares
one depth distribution across all $d$ channels---a forced compromise whose cost
depends on how much independent subspaces disagree about what to read
(probed in Appendix~\ref{sec:mechanism}). The cost is strongly
\emph{data-dependent}: on the anneal corpus a zero-initialized single query
already captures most of the routing gain (Table~\ref{tab:main};
MHAR$-$single-head $= -0.001$/$-0.028$/$-0.006$), whereas in the
web-corpus replication single-head routing degrades from helpful at 100M to
\emph{worse than the plain residual} at 1B while MHAR keeps its full gain
(Appendix Table~\ref{tab:mainweb}). MHAR is thus the best variant in every
cell of both distributions: the multi-head reshape costs nothing and acts as
insurance against the single-query compromise wherever it does bind.
Figure~\ref{fig:perhead} shows the
trained MHAR heads use their per-slice freedom: each maintains a stable,
distinctive deviation from the head-consensus routing---the $H$ heads carry
$H$ different links, not $H$ copies of one.

\paragraph{Downstream evaluation.} To check that the loss improvement is not an
artifact of the training distribution, we evaluate the anneal-trained baseline and MHAR checkpoints
zero-shot (single run) at 100M, 350M, and 1B on held-out WikiText-2 perplexity,
LAMBADA~\citep{paperno2016lambada}, and HellaSwag~\citep{zellers2019hellaswag} with
the LM evaluation harness~\citep{eval-harness} (Table~\ref{tab:downstream}).
Because the models are trained on the anneal mix and the benchmarks are
web-derived, this is a \emph{cross-distribution} transfer test. At
\emph{every} scale MHAR improves perplexity and LAMBADA accuracy, with HellaSwag
better at 350M and 1B and within sampling noise at 100M. The val-loss gain
therefore transfers off the training distribution: the relative perplexity gain
roughly doubles from 100M to 350M and holds at 1B ($-9\%$, $-20\%$, $-19\%$),
and the LAMBADA gain grows monotonically with scale ($+2.4$, $+3.8$, $+7.2$
points); broader benchmarks are future work.

\begin{table}[h]
\centering
\caption{Zero-shot downstream evaluation at 100M, 350M, and 1B (single run) of the
anneal-corpus models of Table~\ref{tab:main}. The benchmarks are web-derived and
held out from the training mix, so gains here are \emph{cross-distribution}
transfer. Each model is evaluated at its training context (100M: seq.\ 2048;
350M/1B: seq.\ 1024), so perplexity is comparable only \emph{within} a scale.
MHAR improves perplexity and LAMBADA at every scale and HellaSwag at 350M/1B
(marginally lower at 100M, within the $\pm3$pt noise of 200 samples).}
\label{tab:downstream}
\begin{tabular}{llccc}
\toprule
\textbf{Scale} & \textbf{Method} & \textbf{WikiText-2 PPL $\downarrow$} & \textbf{LAMBADA $\uparrow$} & \textbf{HellaSwag $\uparrow$} \\
\midrule
100M & Baseline  & 60.0 & 9.6\% & \textbf{35.0\%} \\
100M & MHAR  & \textbf{54.5} & \textbf{12.0\%} & 33.0\% \\
\midrule
350M & Baseline  & 64.9 & 8.8\% & 34.5\% \\
350M & MHAR  & \textbf{52.0} & \textbf{12.6\%} & \textbf{36.0\%} \\
\midrule
1B & Baseline  & 56.4 & 8.8\% & 30.5\% \\
1B & MHAR  & \textbf{45.4} & \textbf{16.0\%} & \textbf{33.0\%} \\
\bottomrule
\end{tabular}
\end{table}

\subsection{Mid-training}
\label{sec:midtrain}

The from-scratch experiments above isolate the routing mechanism at
100M--1B. We now ask whether MHAR survives a realistic 8B \emph{mid-training}
(continual-pretraining) regime, in which an existing base model is annealed on a
capability-oriented corpus. This also stresses the identity-preserving conversion:
we graft multi-head attention residuals onto a pretrained Transformer and cool the
learning rate to convergence.

\paragraph{Data.} We assemble \texttt{anneal\_pt\_v3}, an English-only anneal
corpus of $\approx 1.9$\,T tokens, from public deduplicated, quality-filtered
corpora,
deliberately synthetic-, STEM-, and code-heavy in the style of Nemotron anneal
blends~\citep{su2024nemotroncc}. The blend draws on the Nemotron-CC-v2 web crawl
and its synthetic rephrasing and diverse-QA variants~\citep{nvidia2025nemotronccv2},
the Nemotron pretraining SFT/STEM/code sets~\citep{nvidia2025nemotronsft},
Nemotron-CC-Math~\citep{nvidia2025nemotronccmath}, FinePDFs restricted to its top
three of twenty quality bins~\citep{hf2025finepdfs}, Stack-Edu
code~\citep{hf2025stackedu}, arXiv, and English Wikipedia, shuffled at the
document level into 2{,}048 uniform shards. Every document retains its source
tag, so the composition is measured directly from the shards rather than
estimated: Table~\ref{tab:datamix} (Appendix~\ref{app:repro}) reports per-source
text-byte shares over a uniform random sample of 48 shards.

\paragraph{Base model and schedule.} We start from Marin-8B~\citep{marin2025}
(\texttt{marin-8b-base}, revision \texttt{phoenix}), a Llama-3.1-8B model (8.03B
parameters, $d{=}4096$, $L{=}32$, 32/8 grouped-query heads) released at its
Warmup--Stable--Decay \emph{pre-decay plateau}---the natural resume point for
decay-phase mid-training. The \emph{plain-CPT} control cools from peak
$4\times10^{-4}$ to $4\times10^{-6}$ over 9{,}500 steps at a global batch of
$\approx 1.05$\,M tokens ($\approx 10$\,B tokens total, $\sim0.5\%$ of the
corpus); the MHAR run replicates the control's schedule exactly---the same peak
rate, decay shape, batch, data order, and budget---giving a
\emph{schedule-matched} pair that isolates the routing effect. Optimizer and
systems details (cold-start AdamW with re-warmup, FSDP, EMA, z-loss) are in
Appendix~\ref{app:repro}.

\paragraph{Identity-preserving conversion.} We convert with the \emph{delta}
(additive-stream) form of attention residuals~\citep{luo2026delta}, which
\emph{adds} the routed term to the intact residual stream
($h{=}\mathrm{partial}+\alpha\cdot\mathrm{routed}$) rather than \emph{replacing}
it as the from-scratch block/full form does (Appendix~\ref{sec:kernel}); closing
the gate therefore leaves the pretrained computation untouched. The routed
sources are \emph{block-aggregated}: a learnable null source plus one delta per
block of four consecutive layers (\texttt{num\_blocks}{=}8 for the 32-layer
model), so each routing site attends over at most nine sources with $H{=}8$
heads. MHAR is grafted onto the pretrained model with a zero-initialized output
gate ($\alpha{=}0$ at init), so the converted model is \emph{numerically
identical} to the base at step~0 (maximum logit deviation $0$ in fp32), and the
routing opens during mid-training. This lets us continue-train an existing
checkpoint without a loss spike: on the schedule-matched pair (identical seed
and data order) both runs start from the same loss ($1.8616$ vs $1.8615$ at the
first logged step), and the per-step gap stays $36\times$ below batch noise
across all 9{,}500 steps.

\paragraph{Downstream protocol.} We evaluate six capability-oriented tasks
with the LM Evaluation Harness~\citep{eval-harness}, grouped as \emph{General}
---MMLU~\citep{hendrycks2021mmlu} (57-subject average) and
GPQA~\citep{rein2024gpqa} (main split, 0-shot)---and \emph{Math \& Coding}:
GSM8K~\citep{cobbe2021gsm8k} (5-shot, strict exact match),
MATH~\citep{hendrycks2021mathdataset} (Minerva
4-shot~\citep{lewkowycz2022minerva}), and HumanEval~\citep{chen2021codex}
and MBPP~\citep{austin2021mbpp} (pass@1, greedy). MATH is scored with
the format-robust \texttt{math\_verify} checker rather than the harness's
template-strict extractor: annealed models increasingly answer in
\texttt{\textbackslash boxed\{\}} style while abandoning the few-shot
``Final Answer'' template, which strict extraction mistakes for failure despite
mathematically correct solutions. For generative tasks we report paired
per-item exact McNemar tests between the schedule-matched arms.
Table~\ref{tab:midtrain} reports final (EMA) checkpoints.

\begin{table}[h]
\centering
\caption{Downstream accuracy ($\uparrow$) after 8B mid-training on
\texttt{anneal\_pt\_v3} (final checkpoints, EMA weights). The control and MHAR
columns are \emph{schedule-matched}: identical LR schedule, batch, data order,
and $\approx 10$\,B-token budget.}
\label{tab:midtrain}
\begin{tabular}{clccc}
\toprule
& \textbf{Task} & \textbf{Base} & \textbf{Plain CPT} & \textbf{+MHAR} \\
\midrule
\multirow{2}{*}{\emph{General}}
& MMLU          & 0.554 & 0.643 & \textbf{0.645} \\
& GPQA          & 0.266 & 0.315 & \textbf{0.346} \\
\midrule
\multirow{4}{*}{\emph{Math \& Coding}}
& GSM8K         & 0.190 & 0.470 & \textbf{0.502} \\
& MATH          & 0.053 & \textbf{0.191} & \textbf{0.191} \\
& HumanEval     & 0.122 & 0.409 & \textbf{0.415} \\
& MBPP          & 0.148 & 0.386 & \textbf{0.392} \\
\bottomrule
\end{tabular}
\end{table}

Mid-training on the anneal corpus produces large downstream gains (GSM8K more
than doubles, $19.0\rightarrow47.0$; MATH more than triples,
$5.3\rightarrow19.1$; MMLU rises $+8.9$) while the schedule-matched comparison
isolates what routing itself adds: a real gain of $+3.2$ GSM8K
(123 vs.\ 81 discordant items, paired exact McNemar $p{=}0.004$) and $+3.1$
GPQA (27 vs.\ 13, $p{=}0.038$), with MMLU, MATH, and code statistically
unchanged. The routing gain at this scale is thus modest and concentrated on
GSM8K and GPQA; how it interacts with a more aggressive re-training schedule
is left to future work. We release the mid-trained MHAR model and its
schedule-matched control at
\url{https://huggingface.co/wdlctc/marin-8b-cpt-mhar-delta} and
\url{https://huggingface.co/wdlctc/marin-8b-cpt-plain-lr4e4}.

\subsection{Parameter and compute cost}

\begin{table}[t]
\centering
\caption{\textbf{Training speed and memory.} Baseline Transformer vs.\ MHAR
(\texttt{torch.compile}d reference kernels) vs.\ MHAR with our fused routing
kernels, at the three model settings of Table~\ref{tab:main}. Throughput is
relative to the same-scale baseline; MHAR is parameter-matched ($+0.02\%$);
single-head attention residuals are cost-identical to MHAR by construction.
Protocol, absolute medians, and the full breakdown in
Appendix~\ref{app:cost} (Table~\ref{tab:cost}).}
\label{tab:maincost}
\begin{tabular}{lcccccc}
\toprule
& \multicolumn{3}{c}{\textbf{Throughput} (vs.\ base)} & \multicolumn{3}{c}{\textbf{Peak mem.}\ (GB)} \\
\cmidrule(lr){2-4}\cmidrule(lr){5-7}
\textbf{Method} & 100M & 350M & 1B & 100M & 350M & 1B \\
\midrule
Baseline Transformer & $1.00\times$ & $1.00\times$ & $1.00\times$ & 41.5 & 19.4 & 19.0 \\
MHAR & $0.54\times$ & $0.32\times$ & $0.23\times$ & 52.4 & 40.1 & 47.5 \\
MHAR $+$ fused kernels (ours) & $\mathbf{0.88\times}$ & $\mathbf{0.71\times}$ & $\mathbf{0.55\times}$ & \textbf{42.0} & \textbf{20.0} & \textbf{20.1} \\
\bottomrule
\end{tabular}
\end{table}

\begin{table}[t]
\centering
\caption{\textbf{Routing-operation speedup}, isolating the kernels from
Table~\ref{tab:maincost}'s end-to-end numbers: wall-clock of all routing calls
of one microbatch, forward$+$backward, bf16, single H100; the 8B column is the
mid-training delta variant. End-to-end gains are smaller because routing is
only part of a training step (batch settings and the eager-reference
comparison in Appendix~\ref{app:cost}).}
\label{tab:kernelspeed}
\begin{tabular}{lcccc}
\toprule
\textbf{Routing per microbatch (ms)} & \textbf{100M} & \textbf{350M} & \textbf{1B} & \textbf{8B (delta)} \\
\midrule
\texttt{torch.compile}   & 30.2 & 53.1 & 153.2 & 651 \\
Fused Triton (ours)       & \textbf{5.7} & \textbf{11.4} & \textbf{42.4} & \textbf{323} \\
\midrule
\emph{Speedup}           & $5.3\times$ & $4.7\times$ & $3.6\times$ & $2.0\times$ \\
\bottomrule
\end{tabular}
\end{table}

\paragraph{Parameter cost.} MHAR adds only $+0.02\%$ parameters over the vanilla
baseline (per-layer routing queries and RMSNorms: $+100$K at 350M, $+187$K at
1B; Table~\ref{tab:maincost})---far too few to account for the $-0.06$ to
$-0.15$ nat gain (Table~\ref{tab:main}). Crucially, the
multi-head split itself is \emph{exactly} free over single-head routing in
parameters, compute, \emph{and} memory: $H$ merely reshapes one query into $H$
smaller ones, replacing one softmax with $H$ softmaxes over the short depth axis
(length $\le 2L{+}1$), so single-head and MHAR carry identical throughput and peak
memory by construction (Appendix~\ref{app:cost}). Every MHAR-vs-single-head
comparison in this paper is therefore not merely an iso-parameter, iso-compute
control but iso-\emph{wall-clock}: giving single-head routing more time cannot
close the gap, and the improvement cannot come from added parameters,
normalization, or compute.

\paragraph{Equal-compute comparison.} The one comparison that is \emph{not}
iso-compute is the attention-residual family vs.\ the plain baseline:
materializing and routing over the list of past sublayer outputs has a mechanism
cost, inherited from~\citet{kimi2025attention} rather than introduced by the
multi-head split, so we report the improvement \emph{over the plain Transformer}
at equal steps. Our fused routing kernels (Appendix~\ref{sec:kernel}) remove most
of this cost (Table~\ref{tab:maincost}; the routing operation in isolation speeds
up by $2.0$--$5.3\times$ over \texttt{torch.compile}, Table~\ref{tab:kernelspeed}): MHAR trains at
$0.88\times$/$0.71\times$/$0.55\times$ baseline throughput at 100M/350M/1B with
approximately baseline peak memory, so an equal-wall-clock baseline corresponds to
${\sim}1.1$--$1.8\times$ more steps; a direct iso-compute study against the plain
baseline is future work.

\section{Ablations}
\label{sec:ablation}

\begin{figure}[t]
\centering
\begin{minipage}[c]{0.51\textwidth}
\centering
\includegraphics[width=\linewidth]{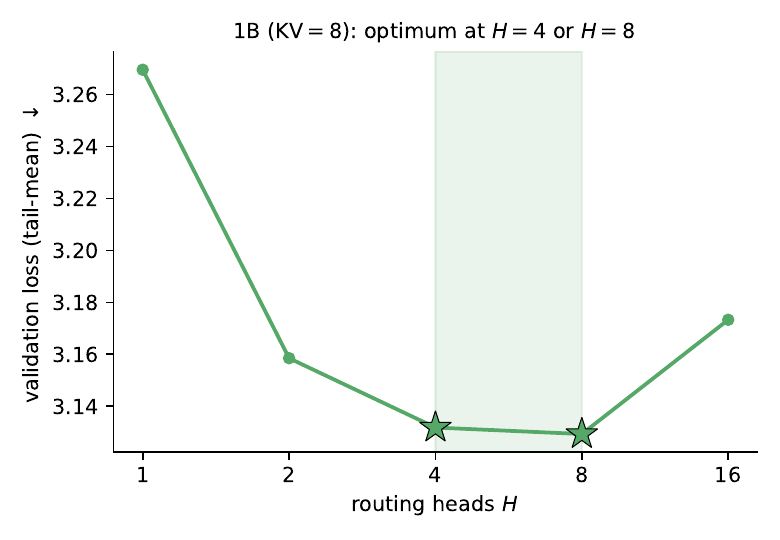}
\end{minipage}\hfill
\begin{minipage}[c]{0.47\textwidth}
\centering
\includegraphics[width=\linewidth]{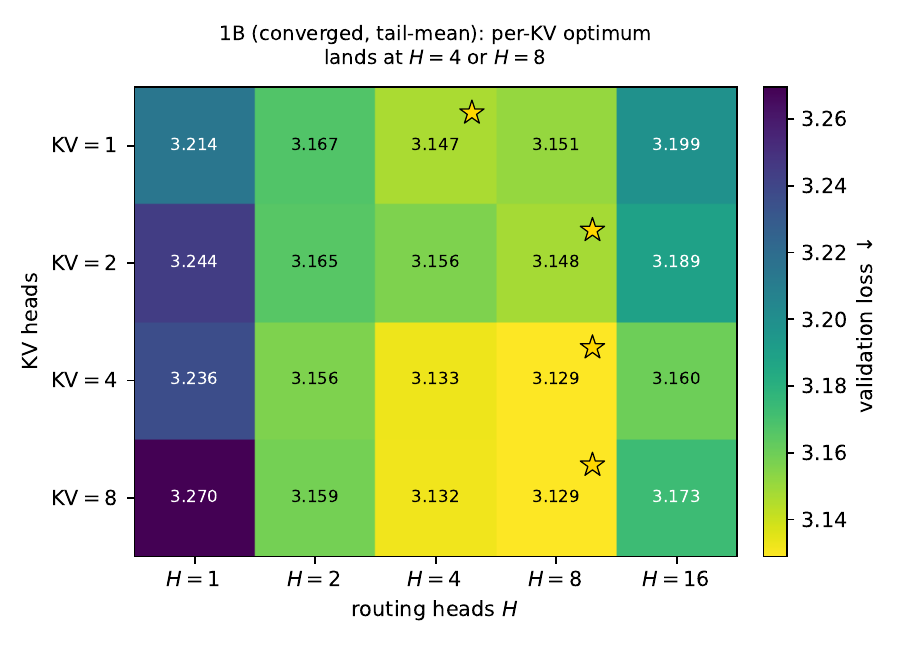}
\end{minipage}
\caption{\textbf{Head count at 1B: a flat $H{=}4$--$8$ optimum.} \textbf{(a)} The
KV$\,{=}\,8$ row (tail-mean): loss falls from
single-head ($H{=}1$) into a flat basin at $H{=}4$--$8$---the optimum, $3.132$
and $3.129$, essentially tied---then rises past it: $H{=}16$ ($3.173$) gives
back roughly a third of the multi-head gain, though it stays well below $H{=}1$.
\textbf{(b)} The full kv$\times$H grid: multi-head routing helps at every KV,
the per-KV optimum (gold stars) lands at $H{=}4$ or $H{=}8$, and the $H{=}16$
column overshoots the basin in every row while staying below $H{=}1$.}
\label{fig:hushape}
\end{figure}

\paragraph{The head count has a two-sided optimum.} Figure~\ref{fig:hushape}a
sweeps the routing-head count at 1B (KV$\,{=}\,8$) with everything else fixed:
loss falls sharply from $H{=}1$ ($3.270$) into a flat basin at the KV
granularity, reaching its optimum at $H{=}4$ ($3.132$) and $H{=}8$ ($3.129$)---
the two are essentially tied, within tail-mean noise---then rises past it:
$H{=}16$ ($3.173$) sits $0.044$ above the optimum, beyond the tail-mean noise
(${\sim}\pm0.02$), and the same overshoot appears in every KV row of the grid
($+0.031$--$0.052$). The two failure modes are granularity mismatches with the
stream's consumers, though asymmetric ones: too few heads force unrelated
subspaces through one depth distribution (costing $0.14$ at $H{=}1$), while too
many split what a KV group consumes coherently (costing $0.03$--$0.05$ at
$H{=}16$, still well below single-head). The size of the overshoot is
corpus-dependent: on the anneal corpus of Table~\ref{tab:main} it is larger
and $H{=}16$ does land above single-head routing. The optimum lands at $H{=}4$ or $H{=}8$ at every
scale we tested---a flat basin---and for large-scale models we adopt $H{=}8$.

\paragraph{Routing should be multi-head, just like attention.}
The attention-residual bypass is structurally a self-attention query over
\emph{depth} (Eq.~\ref{eq:route}) and inherits the same limitation that made
attention multi-head: one query cannot serve every subspace at once
(\S\ref{sec:intro}). The thesis of our work follows: \textbf{if attention is
multi-head, the attention-residual bypass should be multi-head too.} This section
shows that splitting the routing query into $H$ heads is not merely analogous but
the only setting that stays robust as models grow and the data distribution
changes (\S\ref{sec:exp}), and that the optimum consistently sits at $H{=}4$ or
$H{=}8$ (a flat basin, from which we adopt $H{=}8$ at scale), capturing essentially all
of the benefit.

\paragraph{How many heads: a convergence signature.}
How many routing heads help, and where the optimum sits, depends on how converged the
router is. In a deliberately \emph{under-trained} grid (100M on the web corpus
at its tuned LR
$1\times10^{-3}$ but only 5K steps, vs.\ the 20K used elsewhere;
Figure~\ref{fig:heads_under}), the clean signal is in the most over-subscribed row,
KV$\,{=}\,1$: there the single shared query ($H{=}1$) is worst and $H{=}8$ best, a
\emph{monotone} $0.028$ drop, well above the $\sim\!0.006$ single-seed noise floor
(the higher-KV rows are within noise). Before convergence, more heads only ever help.
Trained to convergence, the optimum instead \emph{saturates} into a flat
$H{=}4$--$8$ basin: the full
grids at 100M
(Appendix~\ref{app:kvh100}) and 1B (Figure~\ref{fig:hushape}b) put every per-KV optimum at $H{=}4$ or $H{=}8$
(the over-subscribed KV$\,{\in}\,\{1,2\}$ rows still prefer more heads
even at convergence). We therefore adopt $H{=}8$ for large-scale models (Table~\ref{tab:main}).
This is exactly what the forced-compromise account predicts: the penalty that makes
splitting past KV unhelpful is \emph{learned} (Appendix~\ref{sec:mechanism}) and only
binds once feature subspaces have specialized---before convergence extra heads are
free averaging flexibility; after it, $H{=}4$--$8$ is the knee. Practically,
$H{=}8$ is a safe default for a \emph{well-tuned} run; under-tuning only ever
pushes the optimum to \emph{more} heads, never fewer, so the choice
never costs you.

\paragraph{Single-head routing is not robust across data distributions.}
Single-head routing ($H{=}1$) is exactly the original attention-residual
formulation~\citep{kimi2025attention}: one learned query routes the depth history.
On the anneal corpus it is strong (Table~\ref{tab:main}), but in the
web-corpus replication---read against the baseline with \emph{each method at
its own optimal learning rate} (Appendix Table~\ref{tab:lrsweep})---it is
\emph{not} robust as
the model grows: it helps at
100M ($-0.039$) but lands above the baseline at both larger scales ($+0.038$ at
350M, $+0.105$ at 1B). MHAR, in
contrast, improves at every scale on \emph{both} corpora; on web data
multi-head's advantage over single-head routing
\emph{widens} monotonically with scale: $-0.010$, $-0.118$, $-0.168$ at 100M/350M/1B
(best-vs-best; larger still at a shared rate). The zero-parameter reshape is
what makes attention residuals robust---the central claim of this work.

\section{Conclusion}
Multi-head attention residuals remove the forced-compromise bottleneck of a single
routing query by giving each feature subspace its own depth-routing head, at zero
parameter cost over single-head routing. Trained from scratch from 100M to 1B on a
quality-filtered anneal corpus, MHAR improves over a standard Transformer at every
scale with gains that grow from 100M to the larger scales
($-0.061$/$-0.149$/$-0.140$), and it is the
only variant whose gain also survives web-generic pretraining data, where a single
routing query degrades from helpful to harmful and the multi-head advantage
widens with scale; the gain transfers to held-out perplexity and LAMBADA
accuracy. The optimum
consistently lands at $H{=}4$ or $H{=}8$ across scales, and for large-scale
models we adopt $H{=}8$ (aligning routing to attention head groups adds nothing
measurable; Appendix~\ref{sec:mechanism}). Beyond from-scratch training, an
identity-preserving conversion carries MHAR to 8B mid-training (GSM8K $+3.2$,
paired $p{=}0.004$), and fused routing
kernels cut the mechanism's systems cost to $0.55$--$0.88\times$ baseline throughput
at near-baseline peak memory---together making multi-head depth routing practical
beyond research scale. Understanding \emph{why} the optimum sits at $H{=}4$--$8$, and
what individual routing heads learn to attend to, are natural next steps.

% ---- Broader-impacts paragraph commented out per request; unwrap \iffalse..\fi to restore ----
\iffalse
\paragraph{Broader impacts.} MHAR is a parameter-free architectural change to the
residual stream of decoder-only language models. It inherits the general dual-use and
compute-cost considerations of language-model pretraining and introduces no new data
source, training objective, or deployment surface beyond a standard Transformer; we
foresee no societal risk specific to this modification.
\fi

\bibliographystyle{plainnat}
\bibliography{references}

\appendix

\section{Related Work}
\label{sec:related}

\paragraph{Residual and dense connections.} Residual connections~\citep{he2016deep}
and their gated predecessor, highway networks~\citep{srivastava2015highway},
enable very deep training by providing additive identity paths; \citet{veit2016residual}
characterize residual networks as ensembles of paths of varying length.
DenseNets~\citep{huang2017densely} instead concatenate all previous feature
maps, giving each layer direct access to every earlier one. Our sources list
$\mathcal{S}$ is dense in this spirit, but rather than concatenating (which grows
width) or summing (which is the standard residual stream), we \emph{route} over
the sources with a learned softmax, keeping width fixed while making each past
output individually addressable.

\paragraph{Cross-layer aggregation in Transformers.} Several works move beyond the
single additive stream. DenseFormer~\citep{pagliardini2024denseformer} adds a learned
depth-weighted average of all previous block outputs; MUDDFormer~\citep{wang2024muddformer}
learns multi-way, input-dependent dense connections between layers;
RealFormer~\citep{he2021realformer} carries a residual path on the attention scores.
ReZero~\citep{bachlechner2020rezero} and DeepNet~\citep{wang2022deepnet} instead
rescale the standard residual branch for stable deep training. Attention residuals
differ by replacing the additive stream with an explicit softmax \emph{retrieval}
over past sublayer outputs: unlike DenseFormer's fixed depth-weighted average, the
retrieval is input-adaptive and---crucially for us---\emph{splittable} into
per-subspace heads. Making that retrieval multi-headed at no parameter or FLOP cost
is our contribution.

\paragraph{Routing over layer outputs.} Closest to our setting are attention
residuals~\citep{kimi2025attention}, which introduce a learned attention over the
history of layer outputs. We build directly on this formulation and identify its
single routing query as a bottleneck: splitting it into independent per-subspace
heads (Eq.~\ref{eq:mh}) improves quality while staying parameter- and FLOP-matched.
Hyper-connections~\citep{zhu2024hyperconnections} generalize residual connections
into a learnable width-$n$ set of parallel streams, and
manifold-constrained hyper-connections~\citep{deepseek2025mhc} add geometric
constraints to their mixing; these enrich the \emph{connection topology} between
layers, whereas we keep a single stream and enrich the \emph{routing} that reads the
depth history---our heads act on routing distributions, not on separate streams. MHAR
outperforms hyper-connections at matched recipe across all three scales
(Table~\ref{tab:main}). Concurrent work attacks the \emph{same} bottleneck on a
different axis: routing over per-sublayer \emph{deltas} rather than cumulative states
de-correlates the layer sources~\citep{zhang2026ddl}---targeting the \emph{secondary,
hypothesized} source-collinearity factor of our forced-compromise cost---and
\citet{zhang2026duality} analyze residual-stream structure. We instead split the
routing query across feature subspaces, targeting the \emph{primary} width-disagreement
factor that our probe directly supports; the two fixes are complementary.

\paragraph{Conditional computation and mixtures.} Mixture-of-experts
routing~\citep{shazeer2017moe,fedus2022switch} and mixture-of-depths~\citep{raposo2024mixture}
also use learned routing, but route \emph{tokens} to experts or layers to save or
allocate compute. Our routing is over the \emph{depth axis} of already-computed
sublayer outputs and is applied to every token; it changes what each sublayer
reads, not which tokens are processed.

\paragraph{Multi-head mechanisms and interpretability.} Multi-head
attention~\citep{vaswani2017attention} lets different heads attend to different
token-level patterns; we apply the same intuition one level up, letting different
heads attend to different \emph{layers}. Mechanistic analyses of the residual
stream as a shared communication channel that many components read from and write
to~\citep{elhage2021mathematical}---and evidence that transformer layers act
near-linearly on the stream~\citep{razzhigaev2024linear}---motivate giving different
subspaces independent read access to depth, which is precisely what multi-head
routing provides. Methodologically, validating a
lightweight, parameter-cheap architectural change through systematic from-scratch
experiments across scale follows recent studies of attention modifications such
as gated attention~\citep{qiu2025gated}.

\section{Attention residuals as the single-head route}
\label{app:singlehead}
MHAR changes \emph{only} the routing kernel; the \texttt{forward} pass
(Figure~\ref{fig:code}) is exactly that of attention residuals~\citep{kimi2025attention}.
Figure~\ref{fig:route_single} gives the single-head kernel: one shared query produces one
softmax over depth that all $d$ coordinates read through. Dropping it in for \texttt{route}
in the \emph{identical} forward yields attention residuals exactly; equivalently, it is the
multi-head \texttt{route} of Figure~\ref{fig:code} at $\texttt{self.H}{=}1$ (a single head
of width $d$). This is the \emph{single-head} method reported in every table, so the
MHAR-vs-single-head comparison is a strict iso-parameter, iso-forward control that isolates
the multi-head split.

\begin{figure}[h]
\begin{lstlisting}[style=mhar]
def route(self, sources: list[Tensor], proj: Linear, norm: RMSNorm):
    """Single-head depth routing = attention residuals (Kimi 2025):
       one shared query, one softmax over depth, read by all D coords.
       Equals the MHAR route at H=1."""
    V = torch.stack(sources)                   # [N, B, T, D]
    logits = einsum('d, n b t d -> n b t', proj.weight.view(-1), norm(V))
    return einsum('n b t, n b t d -> b t d', logits.softmax(0), V)
\end{lstlisting}
\caption{\textbf{Single-head route (attention residuals).} A drop-in replacement for the
multi-head \texttt{route} of Figure~\ref{fig:code} that leaves \texttt{forward} unchanged:
a single query yields one depth distribution shared by all $d$ coordinates. This is exactly
MHAR with $H{=}1$, and the single-head baseline in all our tables.}
\label{fig:route_single}
\end{figure}

\section{Routing-head sweep at 100M}
\label{app:kvh100}
\begin{figure}[h]
\centering
\includegraphics[width=0.62\textwidth]{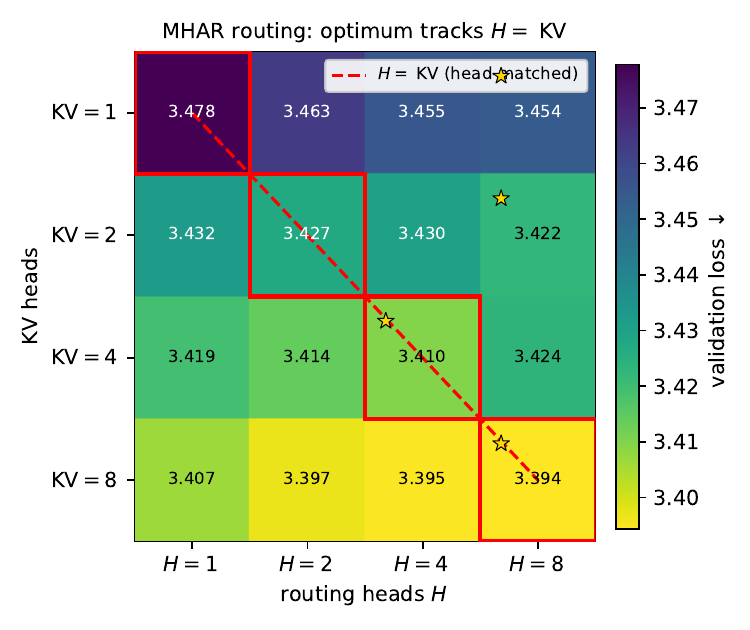}
\caption{Validation loss of MHAR at 100M as a function of routing heads $H$
(x) and KV heads (y). Lower is better. Gold stars mark the best $H$ per KV; the
red dashed line is the head-matched diagonal $H{=}$ KV. At 100M the KV$\,{=}\,4$ and
KV$\,{=}\,8$ optima sit at $H{=}4$ and $H{=}8$ (we adopt $H{=}8$ at scale). The 100M architecture of
Table~\ref{tab:main}, trained on FineWeb-Edu at $5\times10^{-4}$.}
\label{fig:kvh100}
\end{figure}

\begin{figure}[h]
\centering
\includegraphics[width=0.62\textwidth]{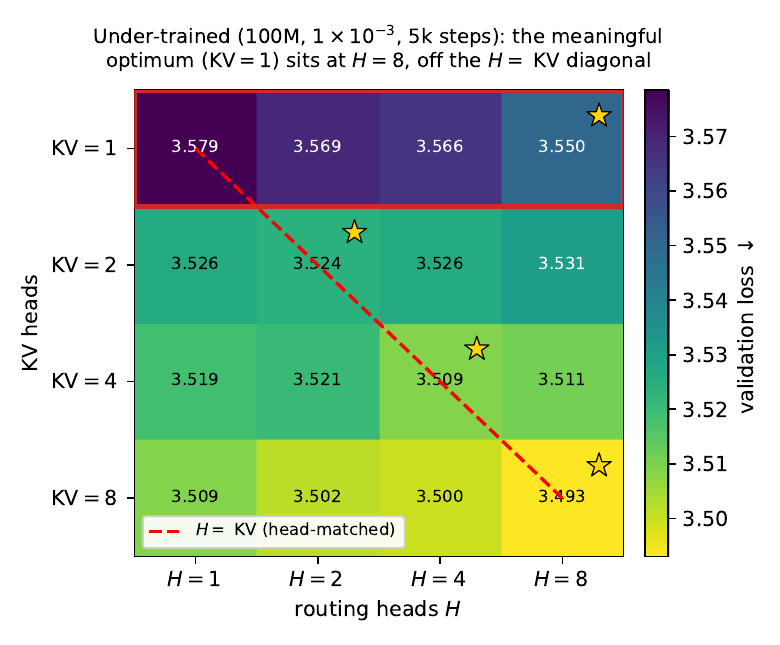}
\caption{\textbf{Under-trained 100M kv$\times$H grid} (web replication,
$1\times10^{-3}$, only 5K steps; tail-mean 3500--5000)---the under-trained
counterpart of Figure~\ref{fig:kvh100}. Before the router converges, more heads
only ever help: the clean supra-noise signal is the most over-subscribed row
KV$\,{=}\,1$ (red outline; a monotone $0.028$ drop from $H{=}1$ to $H{=}8$, above
the $\sim\!0.006$ single-seed noise floor), while the higher-KV rows are within
noise. Gold stars: per-KV optimum; red dashed line: $H{=}$ KV.}
\label{fig:heads_under}
\end{figure}

Figure~\ref{fig:kvh100} shows the full $4\times4$ routing-head sweep at 100M
($H,\text{KV}\,{\in}\,\{1,2,4,8\}$), the converged counterpart of the under-trained
grid in Figure~\ref{fig:heads_under}. These grids train on the web replication
corpus (FineWeb-Edu, randomly initialized routing queries;
Appendix~\ref{app:robustness}), so their absolute losses are not comparable to
the anneal-corpus Table~\ref{tab:main}. At 100M the KV$\,{=}\,4$ and KV$\,{=}\,8$
optima sit at $H{=}4$ and $H{=}8$ (gold stars; on the red $H{=}$ KV line), while at
small KV$\,{\in}\,\{1,2\}$ more heads help, up to $H{=}8$. This
grid pins down the two practically relevant conclusions: multi-head routing helps at
every KV setting, and the optimum consistently lands at $H{=}4$ or $H{=}8$ (we
adopt $H{=}8$ at scale). We ran
the same sweep at 350M and 1B. At 350M the KV$\,{\geq}\,4$ optima again fall at
$H{=}8$. At 1B, however, the global grid argmin is the \emph{off}-diagonal
cell KV$\,{=}\,2$, $H{=}8$ (tail-mean $3.258$), and the four lowest cells---the
KV$2$/$H8$, KV$8$/$H8$, KV$8$/$H4$ and KV$4$/$H8$---all at $H{=}4$ or $H{=}8$---lie
within $\sim\!0.02$ of one another, below the
$\sim\!\pm0.07$ single-pass eval noise; these sweeps are also single-seed. We
therefore read $H{=}8$ at 1B as \emph{within noise of optimal}
rather than the strict grid minimum, and summarize the two sweeps here rather
than as separate figures.

\section{Why single-head routing breaks: the forced-compromise mechanism}
\label{sec:mechanism}

The pattern above has a single explanation. In the web-data setting
(Table~\ref{tab:mainweb}, whose checkpoints this appendix probes), multi-head
routing buys almost nothing
over single-head at 100M---the $H$-axis is flat, MHAR beating single-head by only
$0.010$ (Appendix~\ref{app:kvh100}), while both already beat the baseline, so one
head suffices---yet is worth $\sim\!0.17$ by 1B, where the single query has crossed
from helpful ($-0.039$) to harmful ($+0.105$, best-vs-best). A single $(d,)$ query routes the
entire width through one softmax $\alpha$ (Eq.~\ref{eq:route}), serving every
subspace with one depth distribution. We call the resulting penalty a \emph{forced
compromise}. Its primary, empirically-supported driver is (i) how differently
independent subspaces would route if freed---which grows with model \emph{width}
(more KV subspaces) and which we isolate directly below. A second, hypothesized
factor is (ii) how \emph{collinear} the candidate sources are (near-collinear
sources sharpen the shared softmax and make any disagreement costlier to average
away); this would in principle grow with the cumulative source count
($N{=}2L{+}1$), but we do not cleanly isolate it, and our probe finds it
\emph{non-monotonic} across scale (Table~\ref{tab:probe}). Factor (i) alone already
predicts the observed crossover---and predicts that splitting the query into $H$
heads, each paying the compromise only inside its own $d/H$ slice, removes the
harm. That is exactly what MHAR does.

\paragraph{Direct confirmation on the trained queries.} We measure both factors
directly on the trained single-head ($H{=}1$) checkpoints, with \emph{no} new
training (Table~\ref{tab:probe}). At each routing site we split the trained query
$\mathbf{q}$ into $S$ contiguous slices and compute, for each slice, the depth
distribution it would prefer alone,
$a_s=\mathrm{softmax}_n\langle\mathbf{q}_{[s]},\mathrm{RMSNorm}(\mathbf{s}_n)_{[s]}\rangle$,
relative to the shared $\alpha$ the single softmax actually uses. The mean
$\mathrm{KL}(a_s\,\|\,\alpha)$ is the latent \emph{width-disagreement}. Three
controls make the reading clean.

\emph{(i) The disagreement is learned, not geometric.} A matched-norm \emph{random}
query produces only $\sim\!0.03$--$0.10$ of the disagreement, versus $0.27$--$0.70$
for the trained query, so the signal is overwhelmingly learned rather than an
artifact of source geometry. The learned excess (trained minus null) grows
\emph{monotonically} across all four scales ($0.235\to0.281\to0.512\to0.606$ from
100M to 1B) and is largest at 1B---exactly the scale where single-head routing
\emph{regresses below baseline}, so the probe measures the mechanism directly at the
regression scale. \emph{(ii) Width is the driver.} Widening at \emph{fixed} depth
and KV ($d512\to d768$, $L12$, kv4) raises the trained-query KL by $14\%$
($0.273\to0.311$) while the null stays flat, so the learned excess rises $20\%$
($0.235\to0.281$)---and here $d$ alone changes. \emph{(iii) Not a ``more sources''
artifact.} A matched-source-count control leaves a $\sim\!2\times$ median KL growth,
and contiguous vs.\ head-interleaved slicing give identical results (echoing the
head-alignment null of Table~\ref{tab:headalign}). Source collinearity is \emph{non-monotonic} across scale---it rises from 100M to
350M ($0.060\!\to\!0.114$) but then \emph{drops} at 1B ($0.087$), partly a
dimension effect (cosine shrinks at larger $d$)---so it is \emph{not} a clean
cross-scale driver, and the learned width-disagreement is the signal we rely on.
The forced-compromise cost thus grows with scale and peaks at 1B, tracking the
learned disagreement rather than collinearity, consistent with the loss crossover
in Table~\ref{tab:mainweb}.

Crucially, \textbf{the disagreement is not merely latent}: it moves the loss. At
fixed depth ($L12$, kv4), the same widening $d512\!\to\!d768$ that raises the probed
width-disagreement ($0.235\!\to\!0.281$) also widens MHAR's validation-loss
advantage over single-head routing from $-0.009$ to $-0.020$, while single-head's
edge over the baseline erodes ($-0.041\!\to\!-0.033$)---a matched-recipe training
control that isolates width in loss exactly as the probe isolates it in
disagreement (single-seed; Appendix~\ref{app:widthcost}). \emph{Scope:} the slices
are sub-parts of \emph{one} trained query rather than independently trained routers,
so the KL is an upper-bound proxy for the disagreement of fully independent heads.

The probe measures the disagreement a single query must \emph{suppress};
Figure~\ref{fig:perhead} (main text) shows the complement on the trained 1B MHAR
model: the freed heads maintain stable, mutually distinct deviations from the
head-consensus routing---they route differently, not redundantly.

\begin{table}[h]
\centering
\caption{Direct probe of the trained \emph{single-head} routing queries (no new
training; $S{=}4$ slices). \emph{Width-disagreement} is the mean
$\mathrm{KL}(a_s\,\|\,\alpha)$ between each query slice's preferred depth
distribution and the shared $\alpha$; a matched-norm \emph{random} query gives the
null, and the learned excess (trained$-$null) is the genuine learned signal. The
middle column is a width-isolating control ($d512{\to}d768$ at \emph{fixed} $L12$/kv4/$N$):
the learned disagreement rises while the null and collinearity stay flat, so width
itself drives it. Across scale the learned excess grows monotonically $2.6\times$
($0.235\!\to\!0.606$) and is the \emph{primary, empirically-supported} factor;
source collinearity (mean pairwise cosine of the $N{=}2L{+}1$ sources)---a
\emph{secondary, hypothesized} factor we do not cleanly isolate---is non-monotonic
($1.9\times$ from 100M to 350M, then \emph{lower} at 1B, partly a dimension
effect), so the disagreement factor is the clean cross-scale signal.
\emph{Caveat:} slices are sub-parts of one trained query (a proxy for independent
heads) rather than independently trained routers.}
\label{tab:probe}
\small
\setlength{\tabcolsep}{3pt}
\begin{tabular}{lcccc}
\toprule
 & \textbf{100M} & \textbf{124M} \emph{(width ctrl)} & \textbf{350M} & \textbf{1B} \\
\textbf{Probe quantity ($S{=}4$)} & $d512/L12/$kv4 & $d768/L12/$kv4 & $d1024/L24/$kv8 & $d1280/L36/$kv8 \\
\midrule
Width-disagreement KL (trained) & 0.273 & 0.311 & 0.570 & 0.701 \\
\quad random-query null          & 0.038 & 0.030 & 0.058 & 0.095 \\
\quad \textbf{learned excess} (trained$-$null) & \textbf{0.235} & \textbf{0.281} & \textbf{0.512} & \textbf{0.606} \\
Source collinearity (mean cos)  & 0.060 & 0.063 & 0.114 & 0.087 \\
$N$ sources ($=2L{+}1$)         & 25    & 25    & 49    & 73 \\
\bottomrule
\end{tabular}
\end{table}

\paragraph{Aligning routing to attention head groups.}
A natural hypothesis is that routing should be \emph{aligned} to the model's
attention structure rather than to arbitrary coordinate slices. We test this with
\emph{MHAR-HW}, a head-\emph{wise} variant that gives each grouped-query-attention
KV head group its own full-width routing query and feeds that group's routed
mixture to its $Q/K/V$ projections; with one full-width query per group, its
routing cost is $G\times$ that of MHAR ($G$ the number of KV groups) but still
negligible next to the sublayers. Table~\ref{tab:headalign} compares the two across scales:
head-alignment yields \emph{no detectable} improvement over arbitrary subspace
routing---the two are within the $\pm0.07$ eval-noise floor at all three scales (the
$+0.016$ at 1B is a single final-step eval). Because this ablation is single-seed and
within noise, it does not resolve a difference in either direction: we read it as a
\emph{failure to find} a benefit from head-alignment, not as evidence of equivalence.
It is nonetheless consistent with the benefit of multi-head routing coming from
routing the depth history through multiple independent heads rather than from tying
those heads to the attention head groups.

\begin{table}[h]
\centering
\caption{Head-alignment ablation (FineWeb-Edu; validation loss $\downarrow$, final-step eval):
routing over arbitrary subspaces (MHAR) vs.\ aligned to KV head groups
(MHAR-HW), with matched head count. Aligning to attention heads shows no measurable
benefit over arbitrary subspace routing; all differences fall within the $\pm0.07$
eval-noise floor (single-seed, final-step). Separate matched-run batch (\S\ref{sec:exp}); read the within-table $\Delta$.}
\label{tab:headalign}
\begin{tabular}{lccc}
\toprule
\textbf{Scale} & \textbf{MHAR} & \textbf{MHAR-HW} & \textbf{$\Delta$(hw$-$mh)} \\
\midrule
100M & 3.330 & 3.333 & $+0.003$ \\
350M & 3.225 & 3.223 & $-0.002$ \\
1B   & 3.213 & 3.229 & $+0.016$ \\
\bottomrule
\end{tabular}
\end{table}

\section{Fused routing kernels}
\label{sec:kernel}

Depth routing is memory-bound for two reasons. First, it does almost no
arithmetic per byte: scoring a source row against the query and folding it into
the mix (Eq.~\ref{eq:mh}) is a constant number of operations per element, with
none of the data reuse that lets the attention and MLP matrix multiplies run
compute-bound, so throughput is set by memory bandwidth. Second, the traffic is
inherently quadratic in depth: sublayer $s$ scores and mixes all $N_s{=}s{+}1$
prior states, so a training step must move
$\sum_s N_s\,B\,T\,d = O(L^2\,B\,T\,d)$ of source data no matter how it is
implemented. A reference implementation pays this bound several times over
(per-sublayer stacking of the source list, a materialized normalized key tensor,
separate softmax/mix kernels, thousands of gradient-accumulation launches in
backward) and additionally retains the per-sublayer stacked sources for
autograd, the dominant activation cost of the mechanism (Appendix~\ref{app:cost}). We provide a fused Triton implementation
that reaches (approximately) the bound once per pass: a shared source buffer plus
one kernel per routing call in each direction, with deterministic reductions
throughout (no atomics), so runs are bitwise reproducible.

\begin{figure}[h]
\begin{lstlisting}[style=mhar]
def route_fwd(V, q, g, partial, out, W):  # one program per position p
    m = full([H], -inf); l = zeros([H]); acc = zeros([H, K])
    for n in range(N):                 # single pass over the depth sources
        v = load(V[n, p])              # [H, K] tile; the only source read
        r = rsqrt(mean(v * v) + eps)   # RMS statistic over the full d-row
        khat = round(v * r) * g        # fused RMSNorm; no key tensor made
        logit = sum_k(q * khat)        # per-head scores  [H]
        W[n, p] = logit                # raw logit; normalized after loop
        m_new = maximum(m, logit)      # online softmax: rescale, fold in
        acc = acc * exp(m - m_new) + exp(logit - m_new) * v
        l   = l   * exp(m - m_new) + exp(logit - m_new)
        m = m_new
    out[p] = acc / l + (partial[p] if DELTA else 0)
    for n in range(N):                 # tiny second loop: N*H scalars
        W[n, p] = exp(W[n, p] - m) / l # exact weights, saved for backward
\end{lstlisting}
\caption{\textbf{Fused routing forward.} Online softmax over the depth sources
with the RMSNorm fused in: each source row is read once, nothing is
materialized except the $[N, B, T, H]$ routing weights \texttt{W} (a factor
$d/H$ smaller than the stacked sources a reference path retains for autograd).}
\label{fig:kernel_fwd}
\end{figure}

\begin{figure}[h]
\begin{lstlisting}[style=mhar]
def route_bwd(V, W, dout, dV, dq_part, dg_part):  # one program per pos. p
    do = load(dout[p])                     # [H, K]
    S = zeros([H])
    for n in range(N):                     # pass 1: softmax coupling term
        v = load(V[n, p]); w = W[n, p]
        S += w * sum_k(do * v)
    dq = zeros([H, K]); dg = zeros([H, K])
    for n in range(N):                     # pass 2: all gradients
        v = load(V[n, p]); w = W[n, p]
        dlogit = w * (sum_k(do * v) - S)   # softmax backward  [H]
        r = rsqrt(mean(v * v) + eps)       # norm recomputed, not saved
        khat = round(v * r)
        dq += dlogit * (khat * g)          # query grad, register-resident
        dK = dlogit * q                    # grad wrt normalized key
        dg += dK * khat                    # norm-weight grad
        dkp = dK * g                       # grad wrt v * r
        dv = r * dkp - (r**3 / d) * sum(dkp * v) * v + w * do
        dV[n, p] = dv if FIRST_CALL else dV[n, p] + dv  # in-place accum.
    dq_part[p] = dq; dg_part[p] = dg   # per-position partials, host-summed
\end{lstlisting}
\caption{\textbf{Fused routing backward.} Two passes over the source rows: the
first accumulates the softmax coupling term $S$, the second recomputes the norm
from the buffer (the row is re-read anyway) and emits the closed-form key- and
value-path source gradients, accumulating them across routing calls in place
(safe by autograd's reverse-order execution; asserted at runtime).}
\label{fig:kernel_bwd}
\end{figure}

\paragraph{Shared source buffer.} Each forward pass allocates one
$[2L{+}1, B, T, d]$ buffer and writes every residual state into it exactly once as
it is produced. Routing call $s$ reads rows $[0, N_s)$ directly, eliminating the
per-sublayer \texttt{stack} (an $O(N_s)$ copy per call, $O(L^2\,B\,T\,d)$ in
total).

\paragraph{Forward: online softmax over depth with the norm fused in.} One program
per position $(b,t)$ holds a $[H, d/H]$ accumulator in registers and loops over
the $N$ source rows. For each row it computes the RMS statistic and the per-head
logits $\ell_{n,h} = \langle q_h, \widehat{s}_{n,h}\rangle$ on the fly (fp32
arithmetic, with the normalized key rounded to the storage dtype exactly where the
reference rounds it), and folds the row into the accumulator with the standard
running-max/renormalization update, as in online-softmax attention
kernels~\citep{dao2022flashattention}. The normalized keys are never
materialized, and the only activation saved for backward is the routing-weight
tensor $w \in \mathbb{R}^{N \times B \times T \times H}$---a factor $d/H$ smaller
than the stacked sources a reference path retains. In delta modes the kernel adds
the residual stream to the mix before writing the output.

\paragraph{Backward: one kernel, norms recomputed, gradients accumulated in
place.} Given $\partial\mathcal{L}/\partial\mathrm{out}$ and the saved $w$, the
kernel makes two passes over the source rows. The first accumulates the softmax
coupling term $S_h = \sum_n w_{n,h}\,\langle \partial\mathrm{out}_h,
s_{n,h}\rangle$; the second recomputes the RMS statistic from the buffer (the
row is being re-read anyway, so recomputation is free relative to saving keys),
forms $\partial\ell_{n,h} = w_{n,h}(\langle \partial\mathrm{out}_h,
s_{n,h}\rangle - S_h)$, and emits per-source gradients combining the value path
and the key path through the norm: the contribution added to $\partial s_n$ is
\[
r\,(g \odot \partial\widehat{k}_n)
  - \tfrac{r^3}{d}\,\langle g \odot \partial\widehat{k}_n,\, s_n\rangle\, s_n
  + w_{n}\,\partial\mathrm{out},
\qquad
\partial\widehat{k}_n = \partial\ell_n \cdot q,\;\;
r = \big(\tfrac{1}{d}\lVert s_n\rVert^2 + \varepsilon\big)^{-1/2},
\]
with query and norm-weight gradients accumulated in registers and reduced by a
deterministic two-stage tree. Because every source feeds every later routing
call, a reference backward fans these contributions in through thousands of
small tensor additions; the fused backward instead accumulates them into one
shared fp32 buffer \emph{in place}. This is safe because autograd executes the
routing calls' backwards in strictly reverse forward order---every later call is
a graph descendant of the current call's output through the residual chain---so
when call $s$ runs, rows $[0, N_s)$ already contain all contributions from calls
$s' > s$; the first backward (the final routing) stores directly, and the
ordering is asserted at runtime. Only the row owned by the call's own residual
input is handed back to autograd.

\paragraph{Delta variant~\citep{luo2026delta}.} The mid-training configuration
(\S\ref{sec:midtrain}) routes over a short list---a learnable null source plus at
most \texttt{num\_blocks} block-aggregated deltas---so cross-call buffer sharing
buys little. There we use a self-contained per-call function (per-call stack,
per-call gradients, the kernel adds the residual stream), which composes with
FSDP and activation checkpointing by construction. At the 8B configuration
($d{=}4096$, $H{=}8$, $N{=}9$, per-GPU microbatch $2\times4096$, per-call
checkpointing) routing drops from $651$\,ms (\texttt{torch.compile}) to
$323$\,ms per microbatch ($2.0\times$; $6.4\times$ over the eager reference),
raising end-to-end mid-training throughput by ${\sim}30\%$ on $8\times$H100.
The Table~\ref{tab:kernelspeed} from-scratch columns use the
Table~\ref{tab:main} depth/width at reduced batch so the eager reference also
fits in memory (100M: $B{=}2$, 350M/1B: $B{=}1$; ratios are batch-invariant);
the eager reference kernels are a further $1.5$--$2.5\times$ slower than
\texttt{torch.compile} ($7.9$--$11.5\times$ slower than fused).

\paragraph{Verification.} In fp32 the fused path matches the reference
exactly up to reduction order: on routing chains and full models (with and
without activation checkpointing), losses are identical and every parameter
gradient agrees to $\le 2.5\times10^{-6}$ maximum relative error. In bf16 the two
paths round differently; measured against fp32 ground truth, the fused gradients
are on average \emph{closer} to the truth than the reference's (gradient-error
ratio geometric mean ${\approx}0.5$ across batches; gate-scalar gradients are
cancellation-dominated in both paths). Outputs are bitwise identical across
repeated runs. Measured cost across scales is reported in
Appendix~\ref{app:cost} (Table~\ref{tab:cost}).

\paragraph{Pseudocode.} Figures~\ref{fig:kernel_fwd} and~\ref{fig:kernel_bwd}
give device-level pseudocode for the two kernels. One program (thread block)
handles one position $p=(b,t)$ and holds $[H, K]$ tiles ($K{=}d/H$) in
registers; \texttt{V} is the shared source buffer (rows $[0,N)$; the per-call
stack in delta modes); \texttt{W} holds the routing weights---the only
activation saved for backward. \texttt{round} marks the cast to the storage
dtype at exactly the point the reference RMSNorm rounds, so the fused forward
computes the same function. In the backward, \texttt{dV} is the shared fp32
gradient accumulator: \texttt{FIRST\_CALL} (the final routing call, whose
backward autograd executes first) stores, every earlier call accumulates in
place; in delta modes each call owns a fresh \texttt{dV} and always stores.
Query/norm-weight gradients leave the kernel as per-position partials and are
summed on the host---a deterministic two-stage reduction, no atomics.

\section{Compute and memory cost}
\label{app:cost}
Table~\ref{tab:cost} reports training throughput and peak memory at the three model
settings of Table~\ref{tab:main} on a single H100, for three implementations of the
routing. Three points. First, \textbf{single-head attention residuals and MHAR have
identical cost}: the multi-head split is a parameter-, FLOP-, and memory-free reshape
(\S\ref{sec:method}), so single-head and MHAR share every row of Table~\ref{tab:cost}
by construction ($H{=}1$ multi-head routing reproduces single-head exactly); any
overhead is inherited from attention residuals, not introduced by the multi-head
split. Second, the overhead relative to the additive baseline comes entirely from the
attention-residual \emph{mechanism}: every sublayer routes over the whole list of up
to $2L{+}1$ past sublayer outputs, so a training step moves $O(L^2\,B\,T\,d)$ of
source data, and the reference kernels pay this several times over---per-sublayer
stacking of the source list, a materialized normalized key tensor, separate
softmax/mix kernels, and per-sublayer retention of the stacked sources for the
backward pass, which is what exceeds 80\,GB at 1B. \texttt{torch.compile} fuses the
arithmetic but not the stacking, retention, or backward accumulation. Third, our
\textbf{fused Triton implementation} (design in Appendix~\ref{sec:kernel}) removes
the artifact, yielding $1.7$--$2.4\times$ the \texttt{torch.compile}
path---$0.88\times$/$0.71\times$/$0.55\times$ of baseline throughput at
100M/350M/1B---at approximately baseline peak memory, with training unchanged
(fp32 routing gradients match the reference to $\le 2.5\times10^{-6}$ relative;
bf16 within rounding noise; loss curves coincide). The remaining gap to baseline is
the $O(L^2\,B\,T\,d)$ source re-read inherent to dense depth routing, which the
fused kernels run at roughly HBM bandwidth.

\begin{table}[h]
\centering
\caption{Training compute and memory at the three model settings of
Table~\ref{tab:main} (single H100; per-GPU microbatch as in training: batch 8 /
sequence 2048 at 100M, 4/1024 at 350M, 2/1024 at 1B; identical node and software for
all cells; 120-step runs on a verified-healthy GPU). Each cell is the median of
three steady-state windows (within $\pm 1\%$); peak memory reproduces to the
decimal across sessions, and the 350M compiled/fused cells reproduce in a second
independent session. Throughput is relative to the same-scale baseline (absolute
baseline medians: 113.9k / 45.0k / 19.7k tokens/s). Parameters are matched to the baseline up to the $O(d)$ routing
queries ($+0.02\%$). $^{\dagger}$Single-head attention residuals are cost-identical
to MHAR \emph{by construction} in every row---the head split is a parameter-, FLOP-,
and memory-free reshape (\S\ref{sec:method})---so they are not listed separately.}
\label{tab:cost}
\small
\begin{tabular}{lcccccc}
\toprule
& \multicolumn{2}{c}{\textbf{100M}} & \multicolumn{2}{c}{\textbf{350M}} & \multicolumn{2}{c}{\textbf{1B}} \\
\cmidrule(lr){2-3}\cmidrule(lr){4-5}\cmidrule(lr){6-7}
\textbf{Method}$^{\dagger}$ & thr. & mem.\ (GB) & thr. & mem.\ (GB) & thr. & mem.\ (GB) \\
\midrule
Baseline (additive residual) & $1.00\times$ & 41.5 & $1.00\times$ & 19.4 & $1.00\times$ & 19.0 \\
MHAR, reference kernels      & $0.31\times$ & 68.8 & $0.16\times$ & 70.9 & \multicolumn{2}{c}{OOM ($>$80\,GB)} \\
MHAR, \texttt{torch.compile} & $0.54\times$ & 52.4 & $0.32\times$ & 40.1 & $0.23\times$ & 47.5 \\
MHAR, fused Triton (ours)    & $\mathbf{0.88\times}$ & \textbf{42.0} & $\mathbf{0.71\times}$ & \textbf{20.0} & $\mathbf{0.55\times}$ & \textbf{20.1} \\
\bottomrule
\end{tabular}
\end{table}

\section{Width-control: isolating width in loss}
\label{app:widthcost}
The probe (Table~\ref{tab:probe}, control (ii)) isolates \emph{width} as the driver of
subspace disagreement: widening $d512\!\to\!d768$ at fixed depth/KV raises the learned
disagreement $0.235\!\to\!0.281$ while the random-query null and source collinearity stay
flat. Here we show the same widening also moves the \emph{loss}, closing the link from
disagreement to cost in-distribution (unlike an inference-time intervention, which is
off-distribution). We train all three methods at the wider $d768/L12/$kv4 point
(``124M'') on the web corpus of the replication (FineWeb-Edu;
Table~\ref{tab:mainweb}) under a unified $5\times10^{-4}$ recipe (20K steps,
sequence 2048, global
batch 64, $L{=}12$, kv${=}4$) against a matched $d512$ baseline trained the same way
(the 100M \emph{architecture} of Table~\ref{tab:main}; only $d$, the head count, and
the FFN width change). This control is a self-contained $5\times10^{-4}$ comparison
on the web corpus,
so its $d512$ deltas differ slightly from Table~\ref{tab:mainweb}'s tuned-rate
($1\times10^{-3}$) numbers. Within each width
all three methods share the same data-parallel degree, so the method-vs-baseline
\emph{deltas} cancel world-size effects and are directly comparable across widths
(Table~\ref{tab:widthcost}).

\begin{table}[h]
\centering
\caption{Width-control (FineWeb-Edu, tail-mean validation loss, steps 15--20k). Holding depth
($L{=}12$) and KV ($4$) fixed and only widening $d512\!\to\!d768$: single-head routing's
edge over the baseline \emph{erodes} ($-0.041\!\to\!-0.033$) while MHAR holds
($-0.050\!\to\!-0.053$), so MHAR's advantage over single-head \emph{more than doubles}
($-0.009\!\to\!-0.020$). This is the loss-level counterpart of the width-isolated
disagreement in Table~\ref{tab:probe} (which rises $0.235\!\to\!0.281$ over the same
widening). Single-seed; the single-head deficit shift ($0.008$) is near the per-seed
noise floor, so the robust signal is the widening MHAR$-$single gap.}
\label{tab:widthcost}
\begin{tabular}{lcccc}
\toprule
\textbf{Width} & \textbf{Baseline} & \textbf{Single-head} & \textbf{MHAR} & \textbf{MHAR$-$single} \\
\midrule
100M ($d512$) & 3.462 & 3.421 \,($-0.041$) & 3.412 \,($-0.050$) & $-0.009$ \\
124M ($d768$) & 3.220 & 3.187 \,($-0.033$) & 3.167 \,($-0.053$) & $\mathbf{-0.020}$ \\
\bottomrule
\end{tabular}
\end{table}

\section{Robustness of the from-scratch comparison}
\label{app:robustness}
Every run in this appendix trains on web-generic
FineWeb-Edu~\citep{penedo2024fineweb} with randomly initialized routing
queries (predating the zero-initialization fix; \S\ref{sec:exp}), as do the
head grids of Appendix~\ref{app:kvh100} and the checkpoints probed in
Appendix~\ref{sec:mechanism}.

\paragraph{Web-data comparison (FineWeb-Edu).} Table~\ref{tab:mainweb} repeats
the four-method comparison of Table~\ref{tab:main} on web-generic
FineWeb-Edu~\citep{penedo2024fineweb} at the tuned peak learning rate
$1\times10^{-3}$, with the compute-equivalent gain of MHAR (see
\S\ref{sec:exp}). Two differences from the main-text protocol: these runs
predate the zero-initialization fix (routing queries randomly initialized,
deviating from~\citet{kimi2025attention}; on this corpus zero-init moves the 1B
single-head delta from $+0.082$ to $+0.018$ at $5\times10^{-4}$ but does not
change its sign at the tuned rate), and each method is shown at its tuned rate.
The single-head column reverses relative to the anneal corpus: helpful at 100M,
above baseline at 350M--1B, while MHAR keeps its full gain---the
distribution-dependence discussed in \S\ref{sec:exp}.

\begin{table}[h]
\centering
\caption{Web-data (FineWeb-Edu) validation loss ($\downarrow$, tail-mean) at the
tuned peak learning rate $1\times10^{-3}$; $\Delta$ vs.\ the same-rate baseline;
CEG$_{\mathrm{FLOPs}}$ is the compute-equivalent gain of MHAR over baseline.
Randomly initialized routing queries (pre-fix; see text).}
\label{tab:mainweb}
\begin{tabular}{lcccccc}
\toprule
\textbf{Scale} & \textbf{$d$/$L$/KV} & \textbf{Baseline} & \textbf{Hyper-conn.} & \textbf{Single-head} & \textbf{MHAR} & \textbf{CEG$_{\mathrm{FLOPs}}$} \\
\midrule
100M & 512/12/4  & 3.336 & 3.322 \,($-0.014$) & 3.297 \,($-0.039$) & \textbf{3.287} \,($-0.049$) & $1.27\times$ \\
350M & 1024/24/8 & 3.290 & 3.260 \,($-0.030$) & 3.345 \,($+0.055$) & \textbf{3.210} \,($-0.080$) & $1.49\times$ \\
1B   & 1280/36/8 & 3.175 & 3.172 \,($-0.002$) & 3.315 \,($+0.140$) & \textbf{3.111} \,($-0.063$) & $1.38\times$ \\
\bottomrule
\end{tabular}
\end{table}

\begin{figure}[t]
\centering
\includegraphics[width=0.62\textwidth]{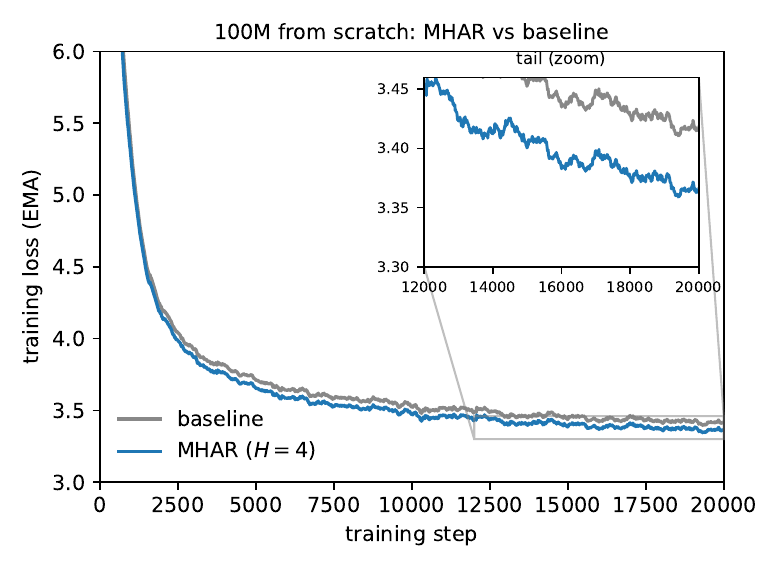}
\caption{Training loss at 100M on FineWeb-Edu (EMA-smoothed): MHAR stays below the standard
baseline throughout training (inset: tail zoom, steps 12k--20k). The two runs are
identical except for the routing mechanism (same node, software, data order, and
global batch).}
\label{fig:trainloss}
\end{figure}

\paragraph{Robustness to learning rate (best-vs-best).} Fixing all methods at
$1\,{\times}\,10^{-3}$ is the optimum for baseline, hyper-connections, and MHAR but
\emph{not} for single-head routing, whose single shared query is destabilized by
the higher rate. To rule out that this either penalizes single-head or favors MHAR, we
sweep the peak learning rate over
$\{1\,{\times}\,10^{-4},\,5\,{\times}\,10^{-4},\,1\,{\times}\,10^{-3}\}$
\emph{independently per method} and compare each at its \emph{own} best
(Table~\ref{tab:lrsweep}). The lowest rate is uniformly too low (validation loss
$4$--$5$: severely under-trained at 20K steps); baseline, hyper-connections, and MHAR
prefer $1\,{\times}\,10^{-3}$; single-head prefers $5\,{\times}\,10^{-4}$ at 350M and
1B (and $1\,{\times}\,10^{-3}$ at 100M). Even
best-vs-best, MHAR improves over the baseline at every scale
($-0.049$/$-0.080$/$-0.063$), and single-head---now at its own kinder
$5\,{\times}\,10^{-4}$---still lands above the baseline at the larger scales
($+0.038$ at 350M, $+0.105$ at 1B,
versus $+0.055$/$+0.140$ at $1\,{\times}\,10^{-3}$). The advantage is a property of
the architecture, not of a favorable or under-tuned learning rate.

\begin{table}[h]
\centering
\caption{Best-vs-best learning rate (FineWeb-Edu): each method at its \emph{own} optimum over peak
LR $\in\{1,5\}{\times}10^{-4},\,10^{-3}$ (validation loss $\downarrow$, tail-mean of
the last eleven evaluations; $\Delta$ vs.\ the same-scale best baseline; superscript
is the selected LR). MHAR wins at every scale even when every method is given its
own optimal learning rate; single-head, even at its kinder $5\times10^{-4}$, still
sits above the baseline at 350M and 1B.}
\label{tab:lrsweep}
\begin{tabular}{lcccc}
\toprule
\textbf{Scale} & \textbf{Baseline} & \textbf{Hyper-conn.} & \textbf{Single-head} & \textbf{MHAR} \\
\midrule
100M & $3.336^{\,10^{-3}}$ & $3.322^{\,10^{-3}}$ \,($-0.014$) & $3.297^{\,10^{-3}}$ \,($-0.039$) & $\mathbf{3.287}^{\,10^{-3}}$ \,($-0.049$) \\
350M & $3.290^{\,10^{-3}}$ & $3.260^{\,10^{-3}}$ \,($-0.030$) & $3.328^{\,5\text{e--}4}$ \,($+0.038$) & $\mathbf{3.210}^{\,10^{-3}}$ \,($-0.080$) \\
1B   & $3.175^{\,10^{-3}}$ & $3.172^{\,10^{-3}}$ \,($-0.002$) & $3.279^{\,5\text{e--}4}$ \,($+0.105$) & $\mathbf{3.111}^{\,10^{-3}}$ \,($-0.063$) \\
\bottomrule
\end{tabular}
\end{table}

\paragraph{Robustness to training budget.} At the fixed 20K-step budget the models
are under-trained, which raises the question of whether MHAR's advantage is
merely an under-training artifact. We therefore train baseline, single-head, and
MHAR for $3\times$ longer (60K steps) at 350M and 1B and read the
baseline-vs-method gap at both budgets (within-run; the longer cosine schedule means
the 20K read here is not comparable to Table~\ref{tab:main}, only the gap-vs-budget
trend is). Tripling the budget lowers validation loss by ${\sim}0.25$---confirming
the runs are genuinely under-trained---but does \emph{not} erase the gaps: MHAR
stays $-0.056$ (350M) / $-0.062$ (1B) ahead of the baseline at 60K (vs.\ $-0.069$ /
$-0.068$ at 20K), and single-head routing's 1B regression persists ($+0.018$ at both
budgets). The architecture effects are thus properties of the model, not transients
of a short schedule.

\paragraph{Seed robustness (all scales).} We train \emph{three seeds per method at
every scale} under a single shared $5\times10^{-4}$ recipe, so that the three
methods differ \emph{only} in seed and routing mechanism. Because seed $s$ fixes the data
order across methods, each method-vs-baseline comparison is \emph{paired},
cancelling cross-seed noise. MHAR improves over the baseline at all three scales,
with the paired delta exceeding ten standard errors everywhere
($-0.045$/$-0.090$/$-0.071$ nats at 100M/350M/1B; Table~\ref{tab:seeds}). The
single-head and hyper-connection patterns also hold under seeds. Single-head routing
moves from a significant \emph{help} at 100M ($-0.043$) through $-0.005$ at 350M
to a significant \emph{regression} at 1B ($+0.093$), and is by far the
highest-variance method: its three 1B paired deltas span $+0.04$ to $+0.16$, versus
MHAR's $\pm0.002$---exactly what the forced-compromise account predicts for an
unstable single shared query. Hyper-connections help at 100M/350M but fall within
noise at 1B ($-0.016$).

\begin{table}[h]
\centering
\caption{Seed robustness (FineWeb-Edu): paired per-seed delta vs.\ baseline (validation loss
$\downarrow$, tail-mean of the last eleven evaluations; three seeds per method per
scale, shared $5\times10^{-4}$ recipe; $\pm$ is the per-seed standard deviation).
Seed $s$ fixes the data order across methods, so the comparison is paired. MHAR
improves at every scale with $|\Delta|/\text{SE}\ge 10$; single-head crosses from
help to harm and is the highest-variance method.}
\label{tab:seeds}
\begin{tabular}{lccc}
\toprule
\textbf{Scale} & \textbf{Hyper-conn.} & \textbf{Single-head} & \textbf{MHAR} \\
\midrule
100M & $-0.031 \pm 0.003$ & $-0.043 \pm 0.003$ & $\mathbf{-0.045 \pm 0.007}$ \\
350M & $-0.028 \pm 0.009$ & $-0.005 \pm 0.015$ & $\mathbf{-0.090 \pm 0.004}$ \\
1B   & $-0.016 \pm 0.015$ & $+0.093 \pm 0.049$ & $\mathbf{-0.071 \pm 0.002}$ \\
\bottomrule
\end{tabular}
\end{table}

\section{Training and reproducibility details}
\label{app:repro}
All runs share a single code path; the settings below are common to every method
and scale unless noted.

\paragraph{Mid-training (8B).} The base checkpoint is released at a constant
peak rate of $1.7\times10^{-3}$ (published before cooldown); our control's peak
$4\times10^{-4}$ is this plateau rate scaled down by
$\sqrt{\text{batch ratio}}$. Because no 8B plateau checkpoint ships loadable
optimizer moments, we cold-start AdamW~\citep{loshchilov2019adamw} and use a
short linear re-warmup ($0\rightarrow$peak) to rebuild the second moment before
decaying. Training uses FSDP full-shard, bfloat16, and activation checkpointing
on $8\times$H100-80GB, sequence length 4096, weight-EMA ($\beta{=}0.999$,
evaluated), a logit z-loss ($10^{-4}$), and weight decay $0.05$ (excluding
embeddings, norms, biases).

\paragraph{Tokenizer and data.} We tokenize the \texttt{anneal\_pt\_v3} corpus
(\S\ref{sec:midtrain}; composition in Table~\ref{tab:datamix}) with the Qwen3
byte-level BPE tokenizer (loaded from \texttt{Qwen/Qwen3-0.6B}). Documents are
streamed from the corpus shards, joined with the end-of-text token (id
$151{,}645$), and
packed into contiguous sequences of length $T$ with no padding. The token-embedding
and output projection are tied (\texttt{tie\_word\_embeddings}) with model vocabulary
size $151{,}936$. The supplementary web replication
(Appendix~\ref{app:robustness}) runs the identical pipeline on
FineWeb-Edu~\citep{penedo2024fineweb}.

\paragraph{Architecture.} Every model follows the Qwen3 recipe~\citep{qwen2025qwen3}:
pre-norm decoder blocks with RMSNorm ($\epsilon{=}10^{-6}$), SwiGLU (SiLU) MLPs,
grouped-query attention with per-head query/key RMSNorm, and rotary position
embeddings (RoPE, $\theta{=}10^{4}$, the training-code default; no learned or absolute
positional embeddings), with $\texttt{max\_position\_embeddings}{=}2T$. Attention
biases are disabled and weights are initialized from $\mathcal{N}(0,0.02^{2})$. There
is no dropout (attention and hidden dropout are $0$).

\paragraph{Optimization.} AdamW~\citep{loshchilov2019adamw} with $\beta_1{=}0.9$,
$\beta_2{=}0.95$, $\epsilon{=}10^{-8}$, and weight decay $0.1$ applied uniformly to
all parameters (no no-decay group). Gradients are clipped to global norm $1.0$. The
learning rate warms up linearly over $1{,}000$ steps to the peak and then follows a
cosine decay to $0.1\times$ peak (the code's default \texttt{lr\_min} ratio) over the
full $20$K-step budget. Training uses bfloat16 (parameters and compute) under
distributed data parallelism at all scales up to 1B.

\paragraph{Batch and sequence length.} The 100M models use sequence length $2048$ and
global batch $64$; the 350M and 1B models use sequence length $1024$ and global batch
$32$. The base random seed is $42$ (offset per worker).

\paragraph{Validation protocol.} There is no explicit held-out split: validation
draws from the \emph{same} corpus under a different shuffle order of the stream (a
$+9999$ seed offset from training, plus a per-worker offset), which reduces rather
than eliminates train/validation overlap. Every $500$ steps we run one evaluation
pass: the token-level cross-entropy is averaged over the pass's evaluation batches and
then averaged (all-reduce mean) across data-parallel workers. Reported numbers are the
\emph{tail-mean} of the last eleven evaluations (steps $15$--$20$K); the single-pass
$\pm0.07$ noise and the paired-comparison rationale are discussed in \S\ref{sec:exp}
(Metric).

\paragraph{Mid-training corpus.} Table~\ref{tab:datamix} gives the measured
composition of the \texttt{anneal\_pt\_v3} corpus used for the 8B mid-training
experiments (\S\ref{sec:midtrain}). The corpus is assembled from public,
quality-filtered corpora, shuffled at the document level into 2{,}048 uniform
parquet shards ($\approx 8$\,TB of text, $\approx 1.9$\,T tokens); every document
carries a \texttt{source} tag and its upstream quality metadata, so shares are
measured, not estimated.

\begin{table}[h]
\centering
\caption{Composition of the \texttt{anneal\_pt\_v3} mid-training corpus: share of
text bytes per source, measured over a uniform random sample of 48 of the 2{,}048
shards ($187$\,GB of text; group rows sum their constituents, so rounded columns
may differ in the last digit). The corpus is English-only; FinePDFs is restricted
to its top three of twenty quality bins (verified from the per-document
\texttt{final\_bucket} metadata).}
\label{tab:datamix}
\footnotesize
\begin{tabular}{lr}
\toprule
\textbf{Source (per-document \texttt{source} tag)} & \textbf{\% bytes} \\
\midrule
\emph{Synthetic web rephrasings (Nemotron-CC-v2/v2.1)} & 24.3 \\
\quad \texttt{nemotron-cc-v2-high-quality-synthetic} & 14.9 \\
\quad \texttt{nemotron-cc-v2.1-high-quality-synthetic} & 6.0 \\
\quad \texttt{nemotron-cc-v2.1-high-quality-translated-to-english-synthetic} & 3.5 \\
\midrule
\emph{Raw high-quality web (Nemotron-CC-v2/v2.1)} & 16.6 \\
\quad \texttt{nemotron-cc-v2-high-quality} & 12.5 \\
\quad \texttt{nemotron-cc-v2.1-high-quality-translated-to-english} & 2.4 \\
\quad \texttt{nemotron-cc-v2.1-high-quality} & 1.6 \\
\midrule
\emph{Web PDFs} & 15.9 \\
\quad \texttt{finepdfs} (top 3/20 quality bins) & 15.9 \\
\midrule
\emph{Synthetic diverse QA (Nemotron-CC-v2/v2.1)} & 14.1 \\
\quad \texttt{nemotron-cc-v2-diverse-qa} & 8.3 \\
\quad \texttt{nemotron-cc-v2-translated-diverse-qa} & 5.3 \\
\quad \texttt{nemotron-cc-v2.1-high-quality-dqa} & 0.4 \\
\midrule
\emph{Code} & 11.9 \\
\quad \texttt{dolma-stack-edu} & 6.7 \\
\quad \texttt{nemotron-pretraining-sft-code} & 2.8 \\
\quad \texttt{nemotron-cc-code-v1} & 2.0 \\
\quad \texttt{nemotron-pretraining-v1.1-code-concepts} & 0.4 \\
\quad \texttt{nemotron-pretraining-scientific-coding} & 0.1 \\
\midrule
\emph{SFT / STEM / reasoning} & 9.7 \\
\quad \texttt{nemotron-pretraining-sft-general} & 4.6 \\
\quad \texttt{nemotron-pretraining-stem-sft} & 4.1 \\
\quad \texttt{nemotron-pretraining-rqa} & 0.7 \\
\quad \texttt{nemotron-pretraining-infinibyte-reasoning} & 0.2 \\
\quad \texttt{nemotron-pretraining-v1.1-multiple-choice} & 0.1 \\
\quad \texttt{nemotron-pretraining-v1.1-formal-logic} & $<$0.1 \\
\quad \texttt{nemotron-pretraining-v1.1-unconditional-algorithmic} & $<$0.1 \\
\midrule
\emph{Math / arXiv} & 6.9 \\
\quad \texttt{nemotron-math-3} & 2.5 \\
\quad \texttt{nemotron-math-4plus} & 1.6 \\
\quad \texttt{nemotron-math-4plus-mind} & 1.1 \\
\quad \texttt{dolma-rpj-proofpile-arxiv} & 0.9 \\
\quad \texttt{nemotron-pretraining-math-textbooks} & 0.8 \\
\quad \texttt{nemotron-pretraining-sft-math} & 0.1 \\
\midrule
\emph{Wikipedia} & 0.6 \\
\quad \texttt{nemotron-pretraining-wiki-rewrite} & 0.3 \\
\quad \texttt{finewiki-en} & 0.3 \\
\bottomrule
\end{tabular}
\end{table}

\end{document}